\documentclass{article}

\usepackage{microtype}
\usepackage{graphicx}
\usepackage{subcaption}
\usepackage{booktabs} 

\usepackage{tabularx}
\usepackage[table, xcdraw]{xcolor}   

\usepackage{amsmath,amsfonts,bm}

\def\eqref#1{equation~\ref{#1}}

\def\1{\bm{1}}

\def\EE{{\textnormal{E}}}
\def\VV{{\textnormal{Var}}}

\def\vx{{\bm{x}}}

\def\vz{{\bm{z}}}

\DeclareMathAlphabet{\mathsfit}{\encodingdefault}{\sfdefault}{m}{sl}
\SetMathAlphabet{\mathsfit}{bold}{\encodingdefault}{\sfdefault}{bx}{n}

\def\cD{{\mathcal{D}}}

\def\cL{{\mathcal{L}}}

\def\cX{{\mathcal{X}}}
\def\cY{{\mathcal{Y}}}

\DeclareMathOperator*{\argmin}{arg\,min}

\def\bbR{{\mathbb{R}}}

\def\bbE{{\mathbb{E}}}

\newcommand{\Err}{\text{Err}}
\usepackage{multicol}
\usepackage{enumitem,kantlipsum} 
\usepackage[colorlinks=true,linkcolor=blue]{hyperref}
\usepackage{url}

\usepackage{multirow}
\usepackage{makecell}
\usepackage{wrapfig}
\usepackage{caption}
\usepackage{rotating}
\usepackage{amsmath}
\usepackage{graphicx}
\usepackage{stackengine,collcell,makecell}
\usepackage{booktabs}       
\usepackage{amsfonts}       
\usepackage{nicefrac}       
\usepackage{microtype}      
\usepackage{bbold}
\usepackage{amsthm}

\usepackage[toc,page,header]{appendix}
\usepackage{minitoc}

\usepackage[normalem]{ulem}

\newcommand{\takeaway}[1]{\textcolor{teal}{\textbf{Takeaway}}: #1}

\usepackage{hyperref}

\usepackage[accepted]{icml2023}

\icmltitlerunning{Synthetic Data, Real Errors: How (Not) to Publish and Use Synthetic Data}

\begin{document}

\twocolumn[
\icmltitle{Synthetic Data, Real Errors: How (Not) to Publish and Use Synthetic Data}



\icmlsetsymbol{equal}{*}

\begin{icmlauthorlist}
\icmlauthor{Boris van Breugel}{cam}
\icmlauthor{Zhaozhi Qian}{cam}
\icmlauthor{Mihaela van der Schaar}{cam,alan}
\end{icmlauthorlist}

\icmlaffiliation{cam}{DAMTP, University of Cambridge}
\icmlaffiliation{alan}{Alan Turing Institute}

\icmlcorrespondingauthor{Boris van Breugel}{bv292@cam.ac.uk}

\icmlkeywords{Machine Learning, ICML}

\vskip 0.3in
]



\printAffiliationsAndNotice{}  

\begin{abstract}
Generating synthetic data through generative models is gaining interest in the ML community and beyond, promising a future where datasets can be tailored to individual needs. Unfortunately, synthetic data is usually not perfect, resulting in potential errors in downstream tasks. In this work we explore how the generative process affects the downstream ML task. We show that the naive synthetic data approach---using synthetic data as if it is real---leads to downstream models and analyses that do not generalize well to real data. As a first step towards better ML in the synthetic data regime, we introduce Deep Generative Ensemble (DGE)---a framework inspired by Deep Ensembles that aims to implicitly approximate the posterior distribution over the generative process model parameters. DGE improves downstream model training, evaluation, and uncertainty quantification, vastly outperforming the naive approach on average. The largest improvements are achieved for minority classes and low-density regions of the original data, for which the generative uncertainty is largest. 
\end{abstract}

\section{Introduction} \label{sec:intro}
Data is the foundation of most science. Recent advances in deep generative modelling have seen a steep rise in methods that aim to replace real data with synthetic data. The general idea is that synthetic data resembles the real data, while guaranteeing privacy \cite{Ho2021DP-GAN:Nets, Yoon2020AnonymizationADS-GAN,Jordon2019PATE-GAN:Guarantees, breugel2023domias}, improving fairness \cite{xu2018fairgan,xu2019achieving,vanBreugel2021DECAF:Networks}, augmenting the dataset size \citep{Antoniou2017DataNetworks, dina2022effect, das2022conditional, Bing2022ConditionalPopulations}, or simulating distributional shifts \cite{Yoon2018RadialGAN:Networks}. Often the aim is to be able to use the synthetic data in place of the real data for some downstream task, e.g. statistical analyses or training an ML supervised model. The hope is that downstream results are equally valid in the real-world---e.g. a prediction model trained on synthetic data will do well on real data. Evidently, whether this is true will rely entirely on how well the synthetic data describes the real data.

This brings us to the focus of this work: how do we do good ML on synthetic data, given that the generative process underlying the synthetic data is not perfect. If we are the data publisher, how should we create and publish synthetic data for it to be most useful to downstream users? And if we are the downstream user, can we create models that are more robust to potential errors, evaluate models reliably using synthetic data, and how do we estimate uncertainty for the downstream results? If we envision a future where synthetic data plays a significant role in research, these are pertinent questions.

\begin{figure}[bt]
    \centering
    \includegraphics[width=\columnwidth]{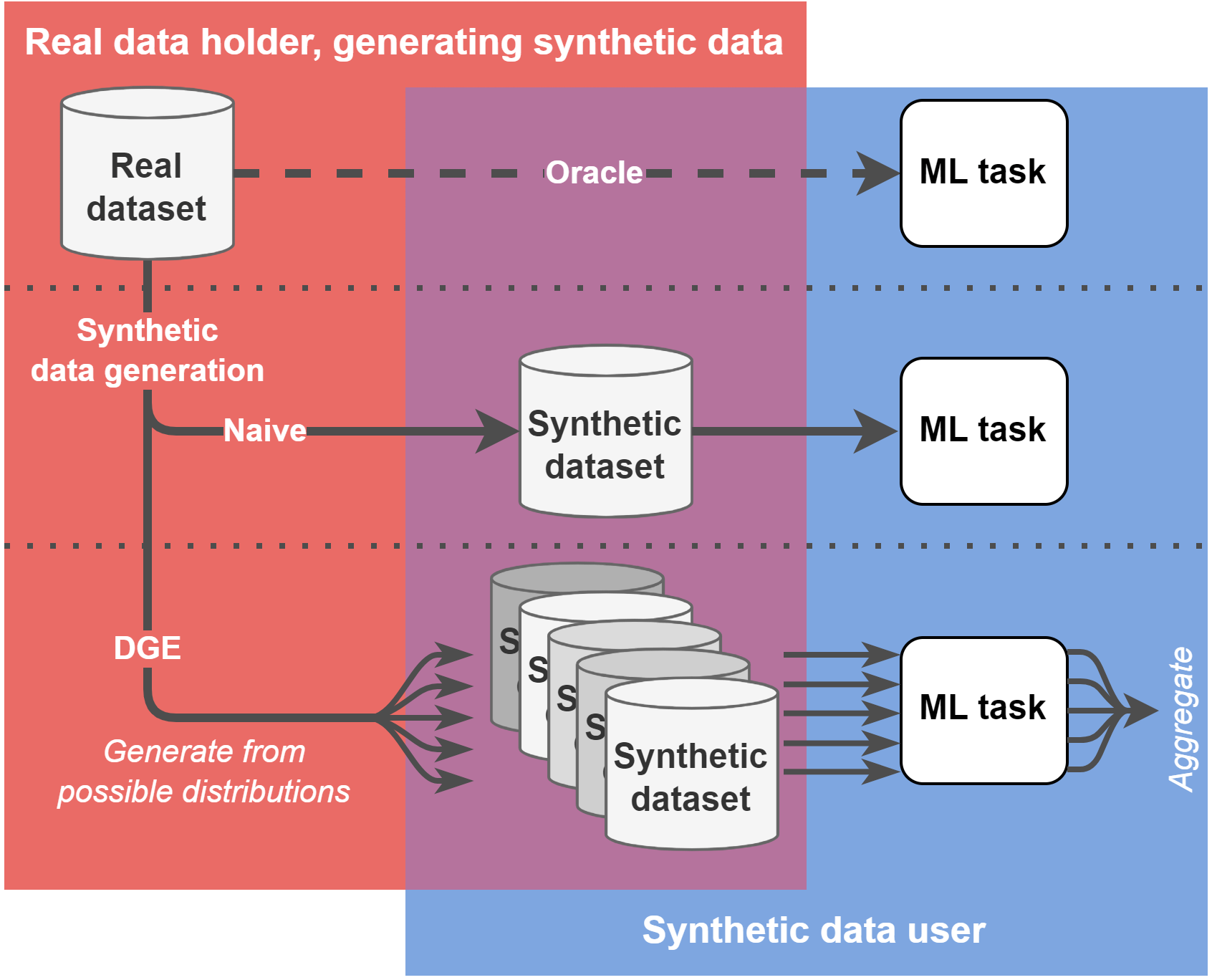}
    \caption{Synthetic data is not perfect, which affects downstream ML tasks, e.g. training a prediction model. The naive synthetic data approach generates one synthetic dataset and treats it like it is real. We propose using an ensemble of generative models for capturing the generative uncertainty, implicitly encoded into different synthetic data distributions.}
    \label{fig:1}
\end{figure}

Let us first highlight why this is an important topic in practice. First, synthetic data is not perfect. Deep generative models may fail in numerable ways, with consequences such as mode collapse, noisy data, memorisation of training data, and poor coverage \cite{breugel2023beyond}. Second, in addition to synthetic data being imperfect, even just quantifying the quality of generative models is hard, because this requires comparing distributions as a whole---a notoriously difficult task \cite{Alaa2022HowModels}. These two points make it hard to guarantee the data is `good' or `bad'.
Third, even if we would be able to measure the quality of synthetic data accurately, in general is not at all trivial how we would use this information for estimating the influence of the generative process on the downstream result---e.g. when training a neural network, the influence of individual training samples is highly complex. Since the downstream user usually has no access to real data (see Figure \ref{fig:1}), they cannot verify results on real data. Let us give a real example of what could you go wrong when using synthetic data.

\begin{wrapfigure}{r}{0.5\columnwidth}
    \centering
    \vspace{-6mm}\hspace{-8mm}
    \includegraphics[width=0.55\columnwidth]{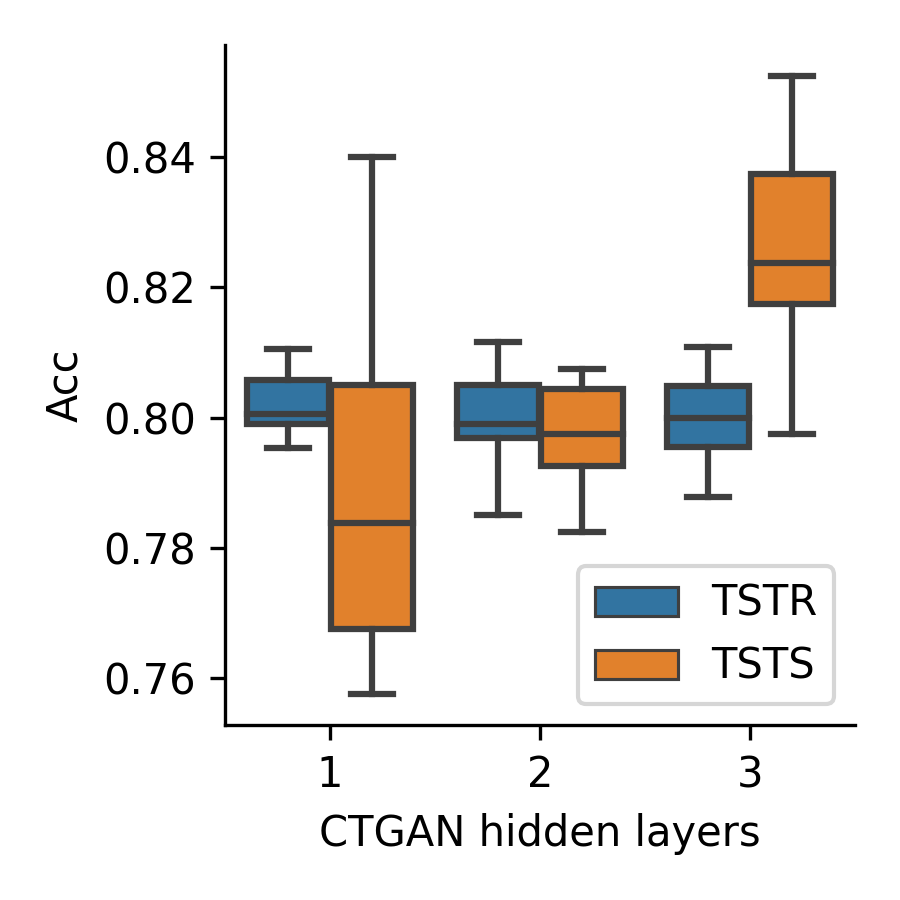}
    \caption{Conclusions drawn from synthetic data do not always transfer to real data.}
    \label{fig:illustrative_example}
\end{wrapfigure}

\textit{Example.} We take the SEER prostate cancer dataset and generate synthetic data using CTGAN with different numbers of hidden layers. Subsequently, we use a train-test-split on the synthetic data and train a random forest for classification. We compare the train-on-synthetic-test-on-synthetic (TSTS) accuracy and the train-on-synthetic-test-on-real (TSTR) accuracy, the latter measured on a real hold-out test set \cite{Jordon2021Hide-and-SeekRe-identification}. Figure \ref{fig:illustrative_example} displays the results over 10 runs. We see that the real performance of the downstream models is comparable across different generative model depths---an indicator that the utility of the synthetic data is similar. On the other hand, the TSTS performance indicates a large preference for the data generated using the deeper generator, with the corresponding TSTS vastly overestimating the TSTR. Also note the TSTS estimates have an undesirable high variance, due to the added randomness in the generative process.

\textbf{Contributions.}
Through this work we explore how---and how \emph{not}--- 
to perform basic ML tasks when using synthetic data. The contributions are as follows.
\begin{enumerate}
 \setlength\itemsep{0em}
    \item We investigate how the standard, naive way of using synthetic data---using it like it is real data---yields poor downstream models, poor downstream model evaluation, and poor uncertainty quantification, due to it ignoring errors in the generation process \emph{itself}.
    \item We introduce Deep Generative Ensemble (DGE) as a simple synthetic data framework for alleviating these concerns through generating multiple synthetic datasets. An advantage of DGE is its flexibility and compatibility with different deep generative base model (e.g. VAE, GAN, diffusion models). 
    \item We investigate why and when DGE provides better downstream  models, evaluation, selection, and better downstream uncertainty quantification.
    \item Furthermore, we explore how DGE improves upon the naive approach especially in low-density regions. This is important, since these regions may correspond to underrepresented groups in the population.
\end{enumerate}
Section \ref{sec:generation} is mostly targeted at data publishers, focusing on synthetic data issues and how DGE aims to fix these. Section \ref{sec:experiments} explores experimentally how the naive approach fails, and describes how DGE-generated synthetic data can be used by data users for better downstream ML. In Section \ref{sec:discussion} we highlight takeaways for both groups.


\begin{table*}[hbt]
    \centering
    \caption{Comparison to related work. (i) Focuses on generative models, (ii) Considers downstream ML tasks, (iii) Considers error in the generative process, (iv) Provides guidelines to synthetic data publishers and users.}
    \scalebox{1}{
    \begin{tabularx}{1.0\textwidth}{lXcccc} \toprule
    Method & Works & (i) & (ii) & (iii) & (iv) \\ \midrule
        Ensembles of generative models & \cite{Tolstikhin2017AdaGAN:Models,Grover2017BoostedModels,Ghosh2017Multi-AgentNetworks,Hoang2018MGAN:Generators}  & $\checkmark$ & $\times$  & $\times$& $\times$\\
        Dropout-GAN & \cite{Mordido2018Dropout-GAN:Discriminators} & $\checkmark$ & $\times$  & $\times$ & $\times$\\
        Deep Ensembles & \cite{Lakshminarayanan2016SimpleEnsembles} & $\times$ & $\checkmark$ & $\times$ & $\times$\\
        MC dropout & \cite{Gal2015DropoutLearning} & $\times$ & $\checkmark$ & $\times$ & $\times$\\ 
        Generative models for UQ & \cite{Bohm2019UncertaintyModels, Phan2019BayesianData, Sensoy2020Uncertainty-AwareModels,Liu2022GFlowOut:Networks} & $\checkmark$ & $\checkmark$ & $\times$ & $\times$\\ \hline
        Deep Generative Ensemble (DGE) &  & $\checkmark$ & $\checkmark$ & $\checkmark$ & $\checkmark$\\ \bottomrule
    \end{tabularx}
    \label{tab:related_work}}
\end{table*}

\section{Related Work} \label{sec:related}
\paragraph{Generative ensembles and dropout.} Deep generative models---notoriously GANs \cite{Goodfellow2014GenerativeNetworks}---often lack diversity in their generated samples, i.e. a GAN that aims to generate patient hospital records may produce a lot of old people but hardly any young people. A large body of work \cite{Tolstikhin2017AdaGAN:Models, Grover2017BoostedModels, Ghosh2017Multi-AgentNetworks, Hoang2018MGAN:Generators} aims to fix diversity issues by ensembling GANs. Some of these approaches use boosting \cite{Tolstikhin2017AdaGAN:Models, Grover2017BoostedModels}, others use multiple generators \cite{Ghosh2017Multi-AgentNetworks, Hoang2018MGAN:Generators}, discriminators \cite{Nguyen2017DualNets, Ghosh2017Multi-AgentNetworks}, or dropout \cite{Mordido2018Dropout-GAN:Discriminators}. These approaches are entirely generative performance focused, and do not consider any application for or insight into improving some downstream ML task. Most importantly, they do not consider how data should be published, and these methods result still in a single synthetic dataset. In Section \ref{sec:experiments} we explore why publishing a single synthetic dataset does not suffice, even if it is generated by an ensemble.

\paragraph{Uncertainty quantification in ML.} Uncertainty quantification has gained significant attention in recent deep learning literature, see \cite{Abdar2021AChallenges} for an overview. One of the more popular methods is Deep Ensembles \cite{Lakshminarayanan2016SimpleEnsembles}, which provides a straightforward approach to supervised model uncertainty estimation: train multiple networks, create predictions using each network and consider the variance between the different networks. Even though this approach is simple, Deep Ensembles have been shown to perform very positively in comparison to fully bayesian methods, possibly due to their ability to capture uncertainty at a more global level of the weight space \cite{Fort2019DeepPerspective}. Note that the other very popular UQ method of Monte Carlo dropout \cite{Gal2015DropoutLearning} can also be seen as an ensemble method, but where the networks share most of the parameters. 
To the best of our knowledge, we are the first to apply UQ to generative models and their downstream task. We note that there are works that consider the opposite: applying generative models to UQ \cite{Bohm2019UncertaintyModels, Phan2019BayesianData, Sensoy2020Uncertainty-AwareModels, Liu2022GFlowOut:Networks}. The aim of these methods is to improve UQ using some form of synthetic data, which is entirely tangential to our work that is interested in the uncertainty in the generative process itself. See Table \ref{tab:related_work} for an overview.

\section{Modelling Generative Uncertainty} \label{sec:generation}
\subsection{Set-up}
Let $X,Y$ be random variables on $\cX, \cY$ denoting features and label, respectively, and let us be given real data $\cD_{r} = (\vx^{(i)}, y^{(i)})_{i=1}^{n_{R}}$ from distribution $p_r(X,Y)$. Let $G_\theta$ be a generator parameterised by $\theta$ that outputs samples with distribution $p_\theta(X,Y)$. We denote samples from $G_\theta$ by $\cD_{s}(\vx^{(i)}, y^{(i)})_{i=1}^{n_{S}}$. 

In the typical generative modelling setting, the objective is to minimise:
\begin{equation}
\label{eq:divergence}
    \theta = \underset{\theta}{\argmin}~ D(p_\theta, p_r)
\end{equation}
for some divergence metric $D$ (e.g. KL-divergence or Wasserstein distance). Though in the limit of infinite training data and capacity some generative models may be guaranteed to achieve $p_\theta = p_r$ (e.g. for GANs \cite{Goodfellow2014GenerativeNetworks}), in practice $p_\theta$ is only an approximation. Evidently, inaccuracies in $p_\theta$ imply that $\cD_s$ has a different distribution as $\cD_r$, and hence this affects any downstream task $T$ we may perform using $\cD_s$. This task $T$ can depend directly on the synthetic data---e.g. estimating the density at some point---or indirectly---e.g. estimating treatment effects or making prediction by first training some supervised ML model $g$. Thus, the variable $T$ is a random variable itself, due to it depending on random $\cD_s$, as well as possible training randomness (e.g. if it is a prediction of a downstream neural network). In any case, we want to take into account the uncertainty in $\theta$ when performing this task.

\subsection{Influence of Data on Downstream Task}
To account for the synthetic data generation process when computing downstream $T$, let us consider the distribution of $T$. Let us denote the distribution of downstream $T$ w.r.t. the real data $\cD_r$ as $p(T|\cD_r)$, we can write:
\begin{equation} \label{eq:T_distribution}
    p(T|\cD_r) = \int p(T|\cD_s)p(\cD_s|\theta)p(\theta|\cD_r) d\cD_s d\theta.
\end{equation}
Let us look at the right-hand-side terms in turn. $p(T|\cD_s)$ is the distribution of $T$ conditional on the synthetic data, which is entirely dependent on the downstream task and something we have no control over as a data publisher. The term $p(\cD_s|\theta)$ is the output of the synthetic data generator for some $\theta$, which we can sample from, but usually have no explicit expression for (in case of deep generative models). At last, $p(\theta|\cD_r)$ is the distribution over the generative modelling parameters given the real data. This is the term we are most interested in; it reflects the uncertainty we have in the generative model parameters themselves.

Computing the integral in Eq. \ref{eq:T_distribution} exactly is intractable for most settings of interest (e.g. if the synthetic data is generated using a GAN). However if a data user would have expressions for all of the terms in Eq. \ref{eq:T_distribution}, they could use Monte Carlo integration for computing any statistic of interest (e.g. the variance). That is, we sample $\hat{\theta}^k\sim p(\theta|\cD_r)$, sample sufficiently large $\cD_s^k\sim p(\cD_s|\theta^k)$, and sample $\hat{T}^{k} \sim p(T|\cD^k_s)$ for $k=1,...,K$, $K\in\mathbb{N}$. This allows us to approximate statistics of interest, for example the empirical mean 
$\EE_{T\sim \hat{p}(T|\cD_r)}[T] = \frac{1}{K} \sum_{k} \hat{T}^{k}$ and variance $\VV_{T\sim \hat{p}(T|\cD_r)}(T) = \frac{1}{K-1} \sum_{k} (\hat{T}^{k}-\EE_{T\sim \hat{p}(T|\cD_r)}[T])^2 $
Evidently, there is a trade-off when choosing $K$: a larger $K$ will give more accurate estimates, but also larger computational costs. We will study this further in the experiments.

\subsection{Modelling the Posterior over $\theta$}
So how do we model $p(\theta|\cD_r)$? The Bayesian approach could motivate us to parameterise the forward generative process, giving $p(\cD_r|\theta)=\prod_i p_\theta(\vz)$, and some prior $p(\theta)$, which would allow computing the posterior over $\theta$:
$
    p(\theta|\cD_r) = \frac{p(\theta)p(\cD_r|\theta)}{\int p(\theta)p(\cD_r|\theta)d\theta}.
$
Computing the denominator is intractable for deep generative models. Consequently, we need to approximate this further. We draw inspiration from the supervised uncertainty quantification (UQ) literature, which aims to estimate $p(\phi| \cD)$ for some predictive model parameterised by $\phi$ trained on data $\cD$. We borrow a popular technique: Deep Ensembles.

\subsection{Approximating the Posterior: Deep Generative Ensemble (DGE)} \label{sec:approximating posterior}

Deep Ensembles \cite{Lakshminarayanan2016SimpleEnsembles} assumes that we can approximate $p(\theta|\cD_r)$ as the empirical distribution over the training process of some deep neural network. In the generative setting, this means we choose a deep generative model class (e.g. VAE, GAN, diffusion model, normalizing flow), train the generative model $K$ times, giving $K$ local solutions $\hat{\theta}^k$ to Eq. \ref{eq:divergence}, and we approximate:
$
    p_{DGE}(\theta|\cD_r) = \frac{1}{K}\sum_k \delta(\theta = \hat{\theta}^k),
$
after which we can use this distribution for computing any downstream statistic of interest. This is a strong assumption and indeed a crude Bayesian approximation of the true posterior---see \cite{Wilson2021DeepInference} for an in-depth discussion. Nonetheless, Deep Ensembles have a solid track record in predictive UQ, often outperforming more traditional Bayesian methods \cite{Fort2019DeepPerspective}. 

\textbf{Choosing the Baselines.} An advantage of DGE is that it allows for different generative model classes. In this paper we focus on tabular data, because many high-stake applications of synthetic data are predominantly tabular, e.g. credit scoring and medical forecasting \cite{borisov2021deep,Shwartz-Ziv2022TabularNeed}. Additionally, nearly 79\% of data scientists work with tabular data on a daily basis, compared to only 14\% who work with modalities such as images \cite{kaggle_2017}. We choose a GAN architecture, specifically CTGAN \cite{Xu2019ModelingGAN}, for its widespread use, and its high expected diversity between individually trained models---cf. VAEs, which tend to learn fuzzier distributions \cite{Theis2015AModels}. We use (i) random initialization for each generative model, and (ii) the same real data for training each base model, as this has been shown to perform well in the supervised domain
\cite{Lakshminarayanan2016SimpleEnsembles}.

\section{Empirical Study: the Effect of DGE on Downstream ML} \label{sec:experiments}

In this section we consider fundamental supervised learning tasks, and how these can be applied in conjunction with synthetic data. We consider using DGE versus the naive synthetic data approach. All experimental details can be found in Appendix A.\footnote{Code and seeded experiments are available at \href{https://github.com/bvanbreugel/deep_generative_ensemble}{{ \texttt{https://github.com/bvanbreugel/\\deep\_generative\_ensemble}}}} 

\subsection{Synthetic Data for Model Training} \label{sec:training}
Let us start by considering how we can train better predictive models on synthetic data. We define ``better'', in terms of a lower generalization error.
Choose some predictive loss function $\mathcal{L}:\bbR^2\rightarrow \bbR$ and predictor $g_\phi:\cX\rightarrow \cY$ parameterised by $\phi$. The generalization error is defined as $
\Err(g_\phi, p_r) = \bbE_{p_r} \mathcal{L}(g_\phi(X), Y).
$
In the case of classification and a proper scoring rule $\cL$, the optimal solution is $g_\phi(x) = p_r(Y|X=x)$. 

Because we do not have data from the real distribution, we cannot minimise the error w.r.t. the real distribution directly. Instead, we aim to choose $\phi$ that minimises: 
\begin{equation}
    \bbE_\theta[\Err(g_\phi, p_\theta)] = \bbE_\theta [\bbE_{(X,Y)\sim p_\theta(X,Y)} L(g_\phi(X),Y))].
\end{equation}
The typical synthetic data approach for model training uses a single synthetic dataset. This will yield high variance in the trained model, because we effectively minimise w.r.t. $p_{\theta_1} (Y|X)$ for a single $\theta_1\sim p(\theta|\cD_r)$. 
Instead, we use an empirical estimate as given by Eq. \ref{eq:T_distribution}: we train a predictive model on each synthetic dataset individually, and average predictions. Let us show how this improves performance. 

\textbf{Datasets.} We use a range of datasets with different characteristics of interest: Scikit-learn's Two-moons and Circles toy datasets---simple two-dimensional datasets that we will later use for visualising the learnt models; UCI's Breast Cancer and Adult Census Income \cite{Asuncion2007UCIRepository}---the former a very small dataset such that synthetic data generation is hard, the latter a large dataset with mixed categorical and numerical features; and SEER \cite{Duggan2016TheRelationship} and Kaggle's Covid-19 dataset \cite{MinistryofHealthofMexico2020Covid-19Dataset}, two large medical datasets with some highly unbalanced categorical features. 

\textbf{Set-up.} We first show that downstream models trained on DGE synthetic data, perform better on real data than baselines. We compare against a classifier trained on a single synthetic dataset (Naive (S)) and 
a pseudo-oracle trained on the real, generative training data ($\cD_r$-model). For fair evaluation, we also include the use of an ensemble of classifiers (Naive (E)) and use the same MLP architecture for all predictive models. At last, we also include a naive generative ensemble approach that concatenates all synthetic datasets (Naive (C)).

We consider the TSTR AUC performance, computed on a hold-out dataset of real data that has not been used for training the generative model. We use CTGAN \cite{Xu2019ModelingGAN} with the same hyperparameters and architecture in all experiments. In Appendix \ref{app:gen_type} we include experiments for other models, showing similar results, and so too do the CelebA and CIFAR-10 results in Appendix \ref{app:image}. See Appendix A for experimental details. 

\textbf{Results.} See Table \ref{tab:model_training}. Training the downstream models on an ensemble of synthetic datasets achieves almost $\cD_r$-model performance on real data. In contrast, the naive baseline is often a few percent lower. This performance increase is \textit{not} merely due to ensembles generally being more robust, since the \textit{Naive (ensemble)} method does not perform as well as DGE$_{20}$, despite using 20 base models. Note that the performance of DGE with $K=20$ is higher on average, but even for $K=5$ we find a significant advantage over the naive (i.e. $K=1$) baseline. 

These results are unsurprising. When the generative model is erroneous---e.g. it overfits or underfits---we expect the naive method to perform poorer than the DGE method, since the different models in the DGE are unlikely to make the same mistakes. 
Inevitably the effect of generative model overfitting is dependent on the downstream task. A simpler downstream task---or simpler downstream model---is less prone to copying the generative overfitting. We will explore this further in Section \ref{sec:selection}.

\begin{table*}[hbt]
    \centering
    \caption{\textbf{Using an ensemble of synthetic datasets for downstream model training improves real-world performance.} AUC performance of different approaches on different datasets, when trained on synthetic data and tested on real data. For the naive methods, we report the median performance across 20 synthetic datasets. Naive (S) uses a single classifier, Naive (E) uses an ensemble of classifiers, though both are trained on a single synthetic dataset. Naive (C) uses all 20 synthetic datasets but naively concatenates them before training a classifier. Note that DGE$_{K}$ gives consistently better performance on average, even for $K=5$.}
    \scalebox{1}{
\begin{tabular}{lllllll|l}
\toprule
{} &          Moons &        Circles &   Adult Income &  Breast Cancer &           SEER &       Covid-19 &   Mean \\
\midrule
$\cD_r$-model              &    0.996 ± 0.0 &    0.868 ± 0.0 &     0.87 ± 0.0 &    0.993 ± 0.0 &    0.907 ± 0.0 &  0.928 ± 0.001 &  0.927 \\
Naive (S)           &  0.981 ± 0.006 &  0.801 ± 0.054 &  0.821 ± 0.006 &  0.975 ± 0.008 &  0.885 ± 0.006 &   0.869 ± 0.02 &  0.889 \\
Naive (E)           &  0.981 ± 0.006 &  0.802 ± 0.053 &  0.837 ± 0.004 &  0.978 ± 0.009 &  0.888 ± 0.006 &  0.895 ± 0.015 &  0.897 \\
Naive (C) &  0.985 ± 0.001 &  0.862 ± 0.005 &  0.852 ± 0.007 &  0.974 ± 0.011 &  0.906 ± 0.001 &  0.895 ± 0.005 &  0.912 \\
DGE$_{5}$           &  0.982 ± 0.002 &  0.853 ± 0.016 &  0.871 ± 0.003 &  0.986 ± 0.003 &  0.903 ± 0.002 &  0.926 ± 0.004 &   0.92 \\
DGE$_{10}$          &  0.983 ± 0.001 &  0.861 ± 0.008 &  0.883 ± 0.002 &  0.986 ± 0.003 &  0.906 ± 0.001 &  0.935 ± 0.003 &  0.926 \\
DGE$_{20}$          &  0.984 ± 0.001 &  0.865 ± 0.003 &  0.889 ± 0.001 &  0.987 ± 0.003 &  0.906 ± 0.001 &  0.942 ± 0.001 &  0.929 \\
\bottomrule
\end{tabular}}
\label{tab:model_training}
\end{table*}

\takeaway{By generating multiple synthetic datasets (e.g. $K=5$), training a prediction model on each, and averaging predictions, one can achieve better performance on real data compared to the naive approach of training a model on a single synthetic dataset. The largest gains are observed when the generative model tends to overfit.}


\begin{table*}[hbt]
    \centering
    \caption{\textbf{Naive synthetic data model evaluation overestimates real-world performance}. Performance of a fixed model, evaluated using different approaches. The naive approach overestimates performance, whereas the DGE approach slightly underestimates it.} 
    \label{tab:model_evaluation}
\scalebox{1}{
\begin{tabular}{llllll|l}
\toprule
{} &          Moons &        Circles &   Adult Income &           SEER &          Covid-19 &   Mean \\
\midrule
Oracle     &   0.775 ± 0.14 &  0.508 ± 0.036 &  0.785 ± 0.015 &  0.711 ± 0.108 &  0.912 ± 0.014 &  0.738 \\ \hline
Naive      &  0.892 ± 0.072 &  0.819 ± 0.132 &  0.784 ± 0.028 &  0.877 ± 0.061 &  0.832 ± 0.042 &  0.841 \\
DGE$_{5}$  &  0.703 ± 0.132 &   0.518 ± 0.07 &   0.773 ± 0.01 &  0.743 ± 0.129 &  0.819 ± 0.022 &  0.711 \\
DGE$_{10}$ &  0.744 ± 0.139 &  0.522 ± 0.094 &   0.774 ± 0.01 &  0.772 ± 0.088 &   0.81 ± 0.017 &  0.724 \\
DGE$_{20}$ &  0.753 ± 0.138 &  0.506 ± 0.045 &   0.775 ± 0.01 &  0.769 ± 0.069 &  0.815 ± 0.016 &  0.724 \\
\bottomrule
\end{tabular}}
\end{table*}

\subsection{Synthetic Data for Model Evaluation and Selection}

Model evaluation and selection (ME/MS) is key to machine learning. ME aims to estimate a model's generalisation error $\Err(g_\phi, p_r)$, while MS aims to choose the model (among a list of models) that minimises this error. 

Estimating the generalization error is usually achieved through train-test splits or cross-validation \cite{Hastie2001TheLearning}. Unfortunately, this approach is not possible when we are in the synthetic data regime \textit{where an ML practitioner has no access to real data}---see Figure \ref{fig:1}. The naive approach is to have a single synthetic dataset and treat it like real data---i.e. one trains a predictive model $f$ on $\cD_{s, train}^k$ and evaluate it on $\cD_{s, test}^k$, for e.g. $k=1$. This induces bias, since the estimate is taken w.r.t. the same generative model. It also has a high variance, because it is an estimate w.r.t. a single draw of $\theta_k$. Can we do better?

As a closer alternative to using an independent real test set---which we do not have---we evaluate w.r.t. other test sets, i.e. $\cup_{i\neq k} \cD_{s, test}^{i}$. This reduces the bias and variance, due to us not using the same model parameters $\theta_k$ for training and evaluation. Let us explore in turn how this influences model evaluation and selection.

\subsubsection{Model Evaluation} \label{sec:evaluation}

\textbf{Set-up.} We split the real data into a training and a test set, and as before train $K$ generative models on the training set, providing $\{\cD_s^k\}_{k=1}^K$. Subsequently, we split up each $\cD_s^k$ into a training and a test set for the downstream model, $\cD_{s, train}^k$ and $\cD_{s,test}^k$. 

We use an MLP as downstream model $g$, trained on a single synthetic training set. We compare the $g$ performance evaluation of the naive approach, our DGE approach and pseudo-oracle evaluation (Oracle)---the latter being the performance of $g$ on a hold-out real dataset. We report results over 20 runs.

\begin{wrapfigure}{r}{0.5\columnwidth}
    \centering
    \vspace{-6mm}
    \hspace{-0.05\columnwidth}\includegraphics[width=0.55\columnwidth]{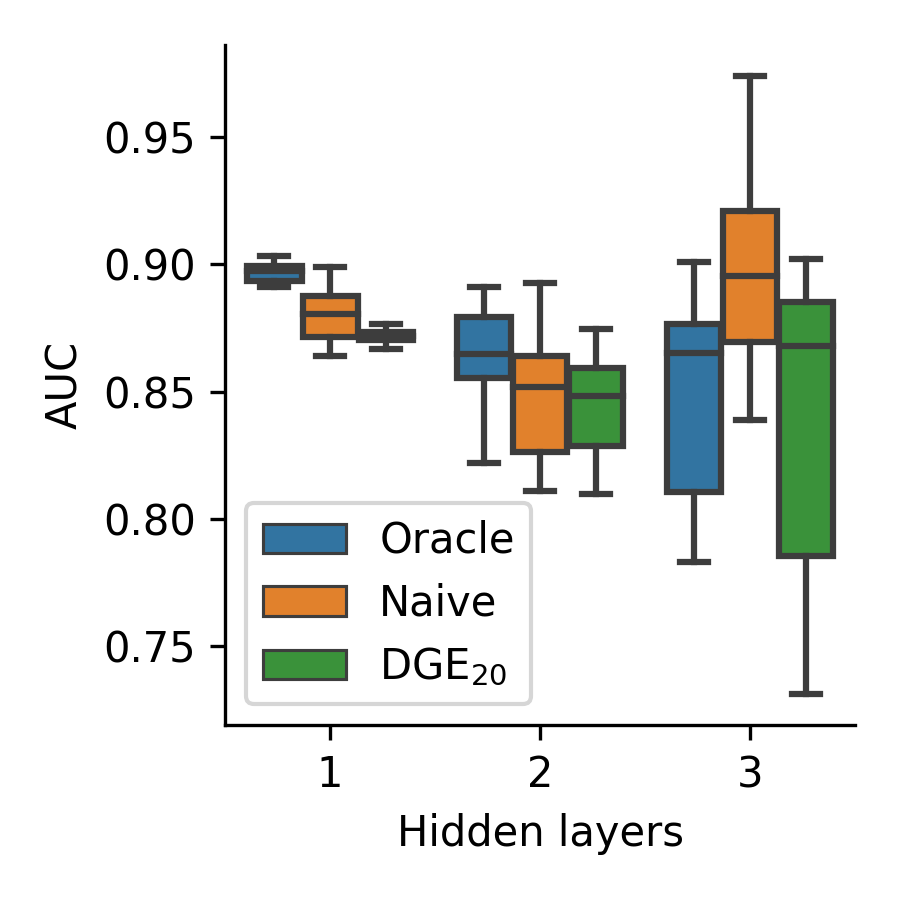}
    \caption{Varying generative size for SEER dataset, shows model evaluation becomes overestimated for the naive approach when the generative model starts overfitting. DGE is more robust to this.}
    \label{fig:story_evaluation}
\end{wrapfigure}

\textbf{Results.} In Table \ref{tab:model_evaluation} we see that the DGE and naive evaluation approaches perform very differently. We see the naive approach overestimates the model performance. This is due to a synthetic data variant of data leakage: overfitting in the generative model is copied by the trained model, but is also reflected in the synthetic test set. DGE evaluation suffers less from this bias, since the test set is from a different generative model, i.e. different draws from $\theta^k$. Conversely, generative errors cause DGE to underestimate the downstream performance often. 
Figure \ref{fig:story_evaluation} shows the same story by varying the generator complexity. An underfitted generative model affects both approaches similarly---slightly underestimated performances. On the other hand, an overfitted generative model leads to significantly overestimated performance by the naive approach. Let us explore how the downstream task plays its part, through considering different downstream models.

\takeaway{Using train-test-splits on single synthetic datasets can lead to significantly overestimated real-world performance, due to train and test split both evaluating on the same (potentially erroneous) generative model. On the other hand, DGE tends to underestimate performance, due to uncertainty in the generative model process leading to different---and aggregated fuzzier---label distributions.}

\subsubsection{Model Selection} \label{sec:selection}


\textbf{Set-up.} In this part, we ask the question: can we decide on which predictive model to use? We use the same set-up as before, but repeat the experiment for the following models: logistic regression, random forest, 5-nearest neighbour, XGBoost, SVM, and a deep MLP---see Appendix \ref{app:details} for experimental details. We consider the ranking of models by different approaches (naive, DGE, and oracle).

\textbf{Results.} We see that DGE ranks models similar to the oracle, whereas the naive approach favours complex models. The latter is again explained by the naive approach's positive bias to a model that captures the generative model's overfitting. Congeniality plays a big role this time; the naive approach is inclined to choose a predictive model that is similar to the generative model's parameterisation. Like most deep generative models, CTGAN uses a neural network architecture for implicitly encoding the label distribution, so it is predictable that this can be best replicated by a deep MLP as downstream model.\footnote{This argument is not entirely straightforward. Since conditional generators like CTGAN usually model the conditional feature distribution $p(X|Y=y)$ using an NN, it is not necessarily true that the output $p(Y|X)$ itself falls in the same model class as the generator. Nonetheless, we do expect the generator output to show similar behaviour (e.g. ReLu artifacts) as the underlying NN, since $p(Y|X) = p(X|Y)p(Y)/(\sum_y p(X|Y=y)p(Y=y))$.} The naive approach becomes even worse when the amount of synthetic data increases, since for more complex models this allows better learning of the generative label distribution---see Appendix \ref{app:synthetic_dataset_size} for experiments.

\begin{table*}[hbt]
    \centering
    \caption{\textbf{Naive evaluation does not preserve real-world model ranking, whereas DGE does.} Model selection on SEER, though evaluating different model classes trained on a downstream synthetic dataset. We see that ranking of models is preserved in all three DGE approaches, in contrast to to the naive approach. Results show mean AUC and standard deviation, taken across 20 runs.}
    \label{tab:model_selection}
\scalebox{0.75}{
\begin{tabular}{llllllll|rrrrrrr}
\toprule
{} &       deep MLP &            SVM &            kNN &        XGBoost &             RF &            MLP & Log. reg. & \multicolumn{7}{c}{Ranking} \\
\midrule
Oracle     &  0.767 ± 0.158 &   0.781 ± 0.16 &    0.8 ± 0.102 &  0.838 ± 0.054 &  0.844 ± 0.053 &  0.849 ± 0.116 &  0.869 ± 0.062 &  7 &  6 &  5 &  4 &  3 &  2 &  1 \\
Naive      &  0.849 ± 0.113 &  0.804 ± 0.187 &  0.823 ± 0.118 &  0.885 ± 0.085 &  0.873 ± 0.075 &  0.867 ± 0.143 &  0.866 ± 0.104 &  5 &  7 &  6 &  1 &  2 &  3 &  4 \\
DGE$_5$  &  0.797 ± 0.189 &  0.819 ± 0.184 &  0.831 ± 0.132 &   0.859 ± 0.07 &   0.87 ± 0.072 &  0.873 ± 0.137 &  0.884 ± 0.088 &  7 &  6 &  5 &  4 &  3 &  2 &  1 \\
DGE$_{10}$ &  0.798 ± 0.144 &  0.811 ± 0.171 &   0.826 ± 0.12 &   0.85 ± 0.064 &  0.861 ± 0.063 &  0.871 ± 0.101 &  0.878 ± 0.065 &  7 &  6 &  5 &  4 &  3 &  2 &  1 \\
DGE$_{20}$ &  0.778 ± 0.139 &  0.796 ± 0.159 &   0.807 ± 0.11 &  0.836 ± 0.064 &  0.849 ± 0.057 &  0.856 ± 0.104 &  0.865 ± 0.071 &  7 &  6 &  5 &  4 &  3 &  2 &  1 \\
\bottomrule
\end{tabular}}
\end{table*}

\takeaway{The naive approach consistently selects more complex models that can copy the generative model's $p_\theta(Y|X)$, but which generalise poorly to real data. DGE has a more accurate estimate of real-world performance due to evaluating on other synthetic datasets, which leads to it selecting simpler downstream models that generalise better to real-world data.}

\subsection{Model Uncertainty}
\label{sec:uncertainty}
Going one step further than evaluation, we look at downstream predictive uncertainty quantification. Using synthetic data like it is real does not account for uncertainty in the generative process itself, which leads to underestimated downstream uncertainty. We focus on classification and define uncertainty in terms of the estimated probability of the predicted label. We show the difference between generative and predictive uncertainty.

\textbf{Set-up.} To separate generative and predictive uncertainty, we include a Deep Ensembles \emph{predictive} model \cite{Lakshminarayanan2016SimpleEnsembles} that provides uncertainty on the predictive level. Effectively, we compare the sample mean and variance in $\hat{P}(Y=1|x)$ of (i) a DGE approach, in which each synthetic dataset is used for training a predictive model, (ii) a naive approach, in which one single synthetic dataset is used to train $K$ predictive models in a Deep Ensembles fashion, (iii) Naive (C), in which all synthetic datasets are concatenated and a Deep Ensembles is trained on this dataset. We add toy dataset Gaussian for visualization---see Appendix A for details---and remove the Breast cancer dataset due to insufficient number of real test samples for accurate assessment.

\begin{figure}[bt]
    \centering
    \begin{subfigure}[b]{0.32\columnwidth}
    \includegraphics[width=\textwidth]{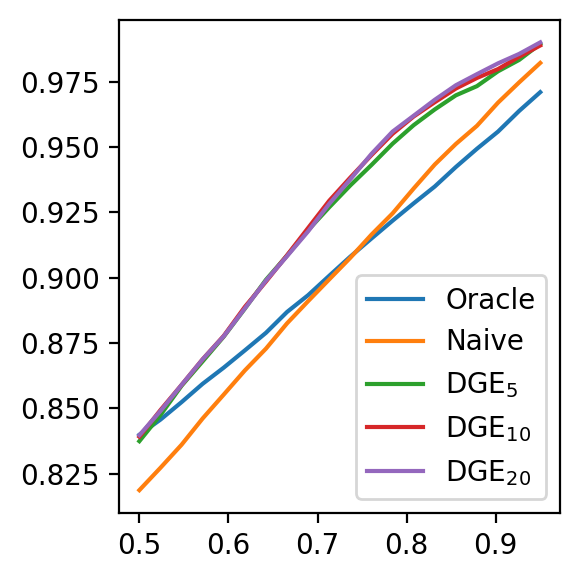}
    \caption{Adult}
    \end{subfigure}
    \begin{subfigure}[b]{0.32\columnwidth}
    \includegraphics[width=\textwidth]{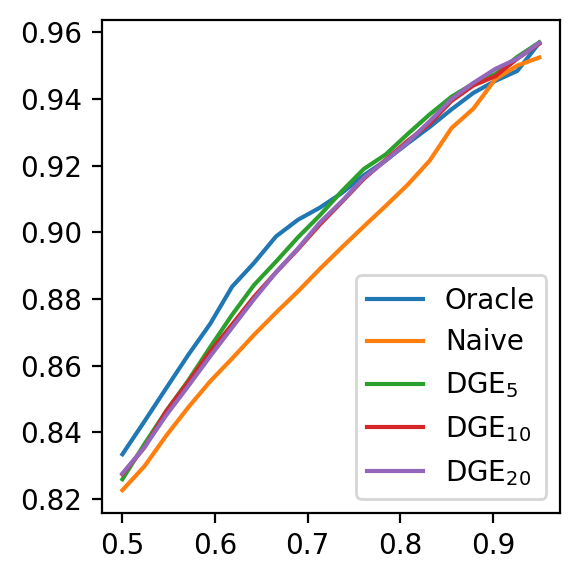}
    \caption{SEER}
    \end{subfigure}
    \begin{subfigure}[b]{0.32\columnwidth}
    \includegraphics[width=\textwidth]{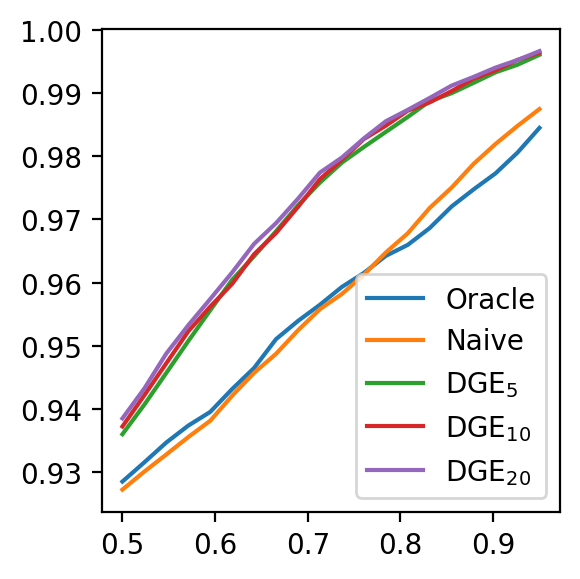}
    \caption{Covid-19}
    \end{subfigure}    
    \caption{Confidence accuracy curves. Given threshold $\tau$ (x-axis), the y-axis shows the accuracy on test sample with confidence $\hat{P}(Y=\hat{y}|x)>\tau$. DGE achieves consistently higher accuracy for different confidence thresholds.}
    \label{fig:confidence_accuracy}
\end{figure}

\begin{figure}[hbt]
    \centering
    \setlength{\tabcolsep}{0pt}
\begin{tabular}{clll} 
  &
  \makecell[c]{Two-moons} &
  \makecell[c]{Circles} &
  \makecell[c]{Gaussian} 
  \\ \midrule
\rotatebox{90}{\quad Samples} &
  \includegraphics[height=0.12\textwidth, trim = 0 3mm 0 0, clip]{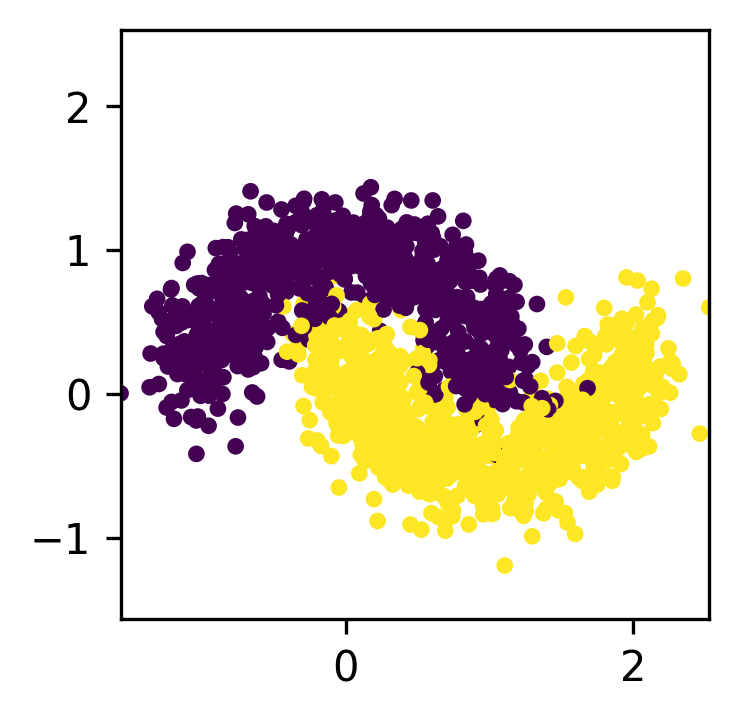} &
  \includegraphics[height=0.12\textwidth, trim = 0 3mm 0 0, clip]{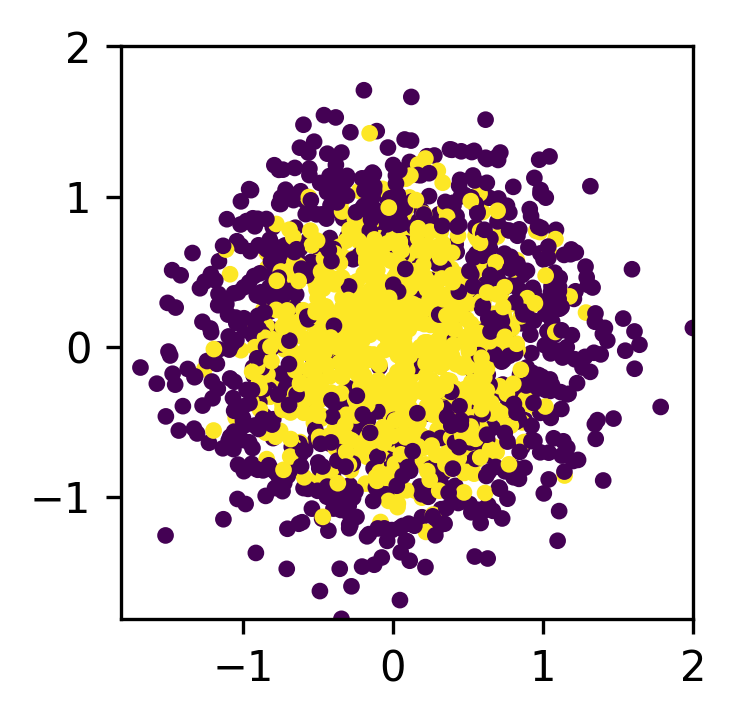} &
  \includegraphics[height=0.12\textwidth, trim = 0 3mm 0 0, clip]{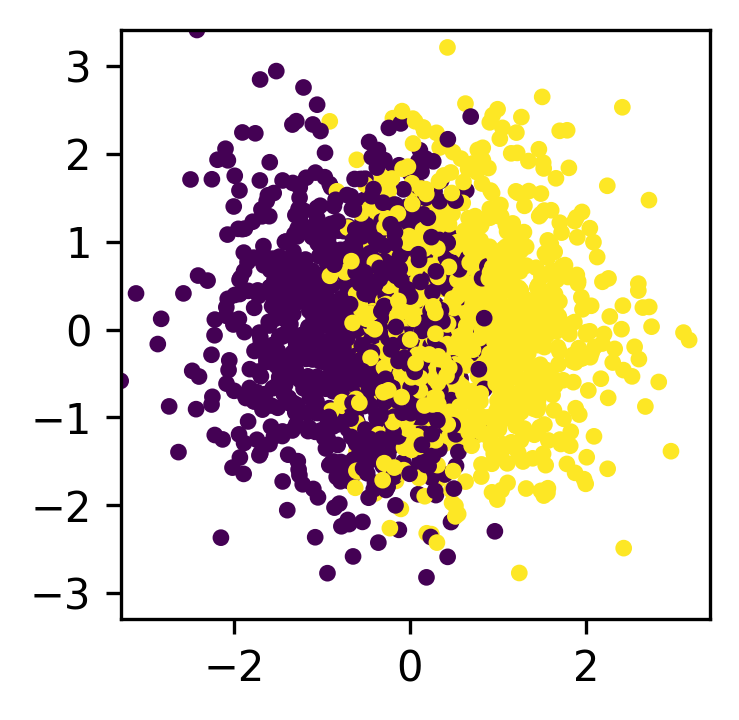} \\ \hline
\rotatebox{90}{\ \ $\cD_r$-model} &
  \includegraphics[height=0.12\textwidth, trim = 0 3mm 0 0, clip]{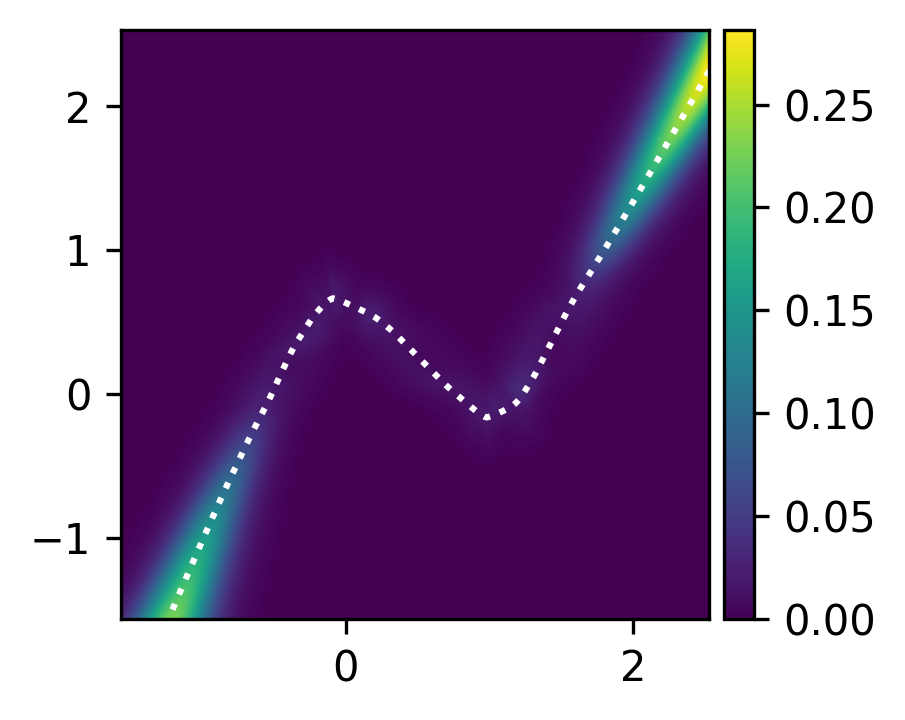} &
  \includegraphics[height=0.12\textwidth, trim = 0 3mm 0 0, clip]{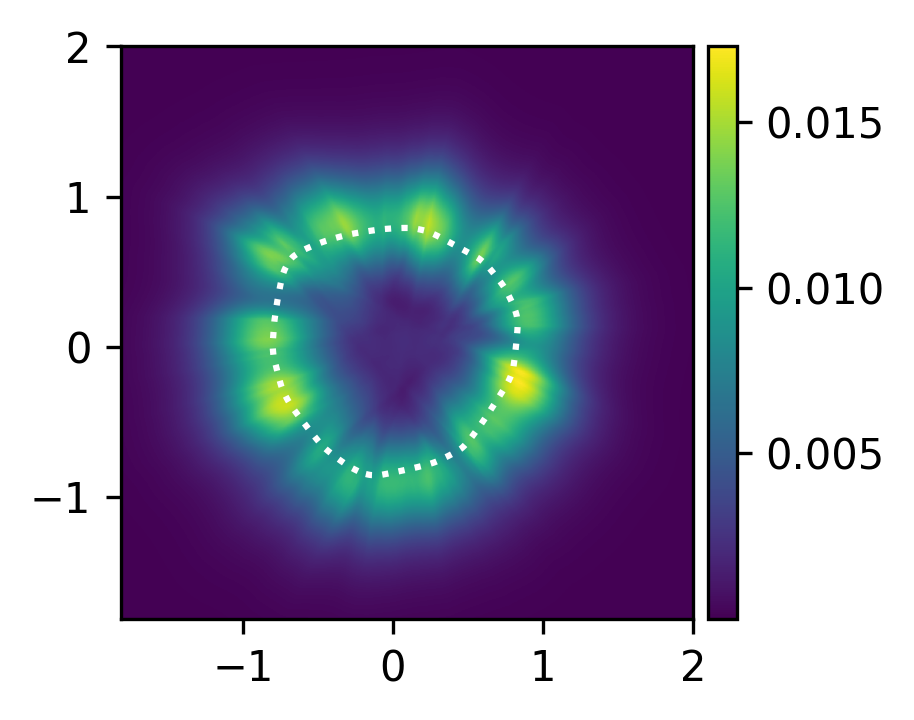} &
  \includegraphics[height=0.12\textwidth, trim = 0 3mm 0 0, clip]{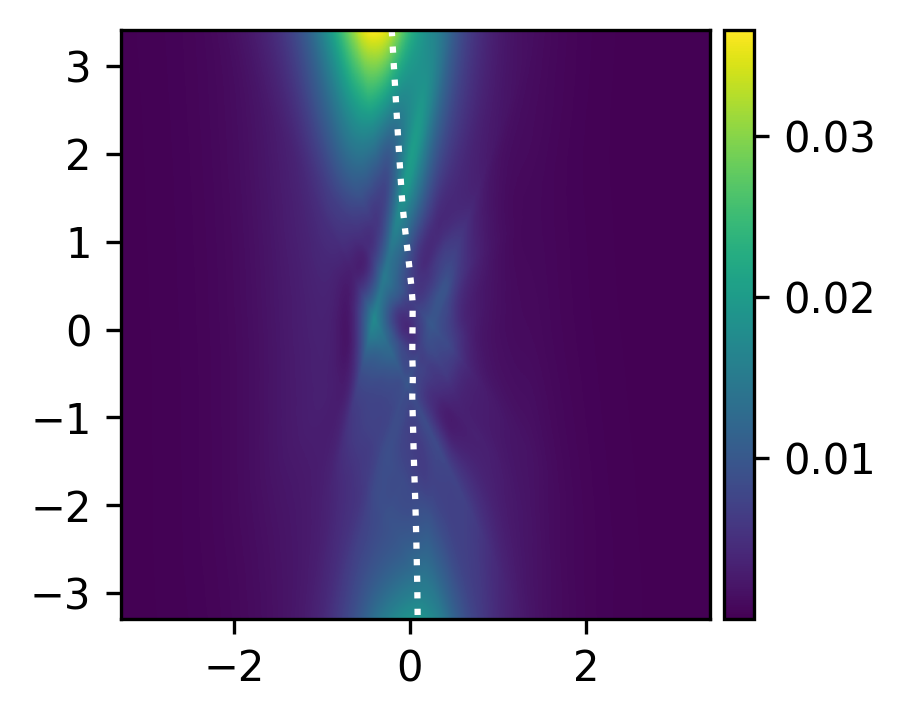} \\ \hline
\rotatebox{90}{\quad \ \ Naive} &
  \includegraphics[height=0.12\textwidth, trim = 0 3mm 0 0, clip]{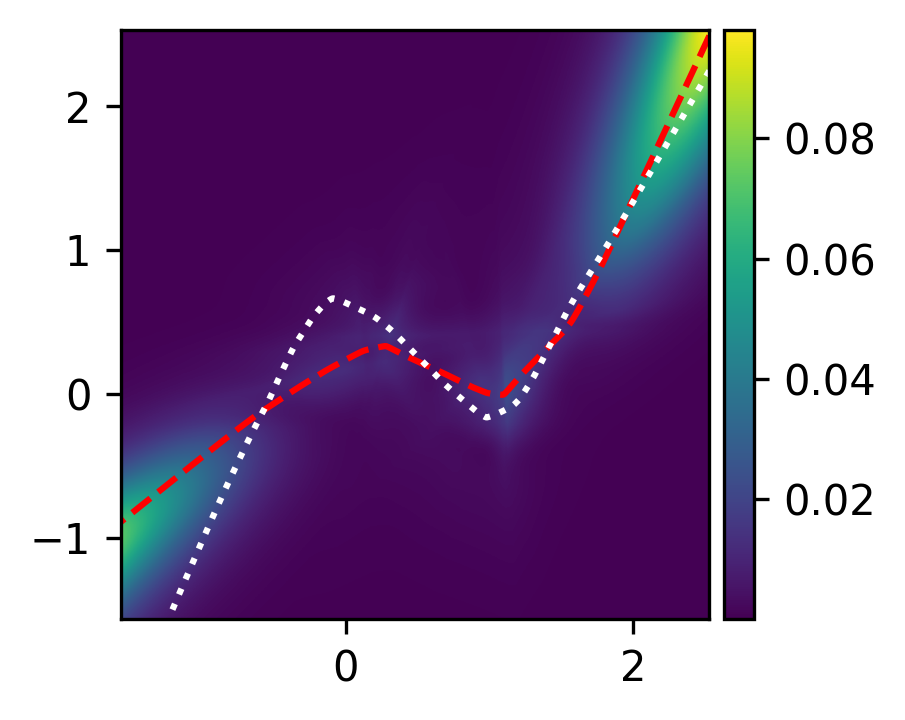} &
  \includegraphics[height=0.12\textwidth, trim = 0 3mm 0 0, clip]{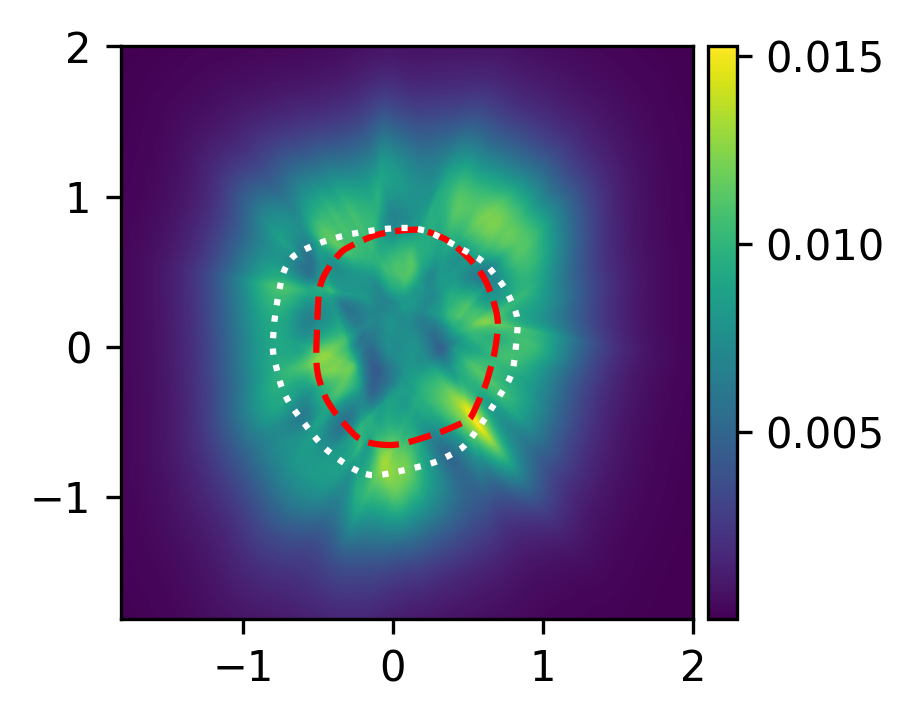} &
  \includegraphics[height=0.12\textwidth, trim = 0 3mm 0 0, clip]{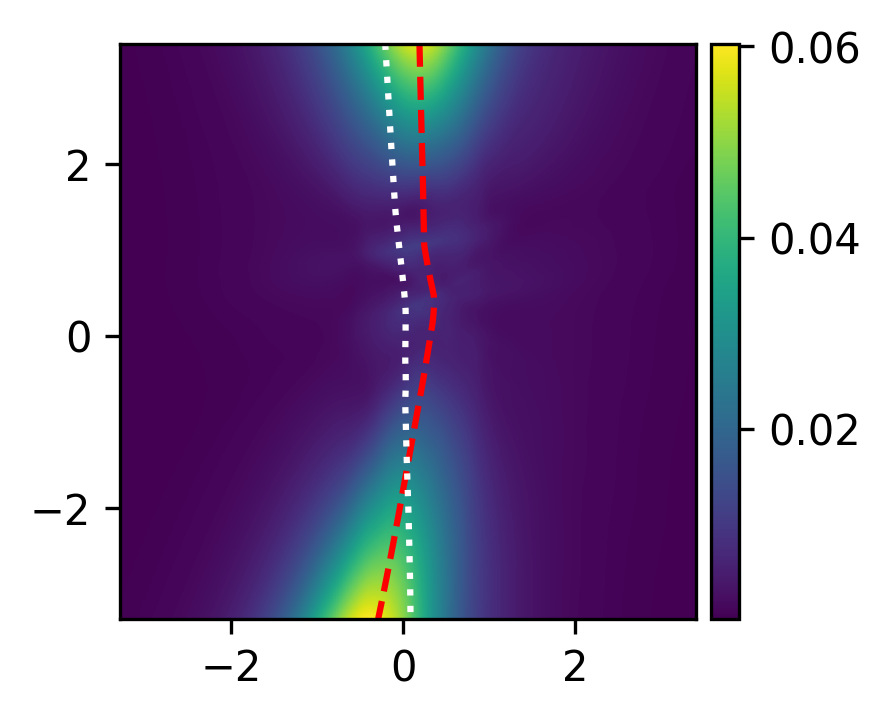} \\ \hline
\rotatebox{90}{\quad \ Naive (C)} &
  \includegraphics[height=0.12\textwidth, trim = 0 3mm 0 0, clip]{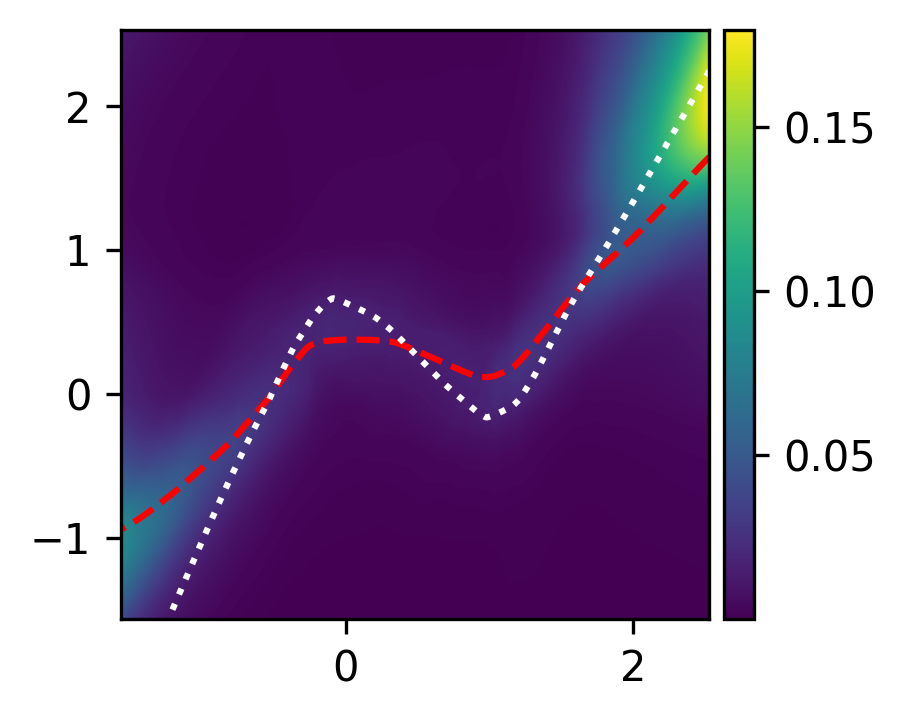} &
  \includegraphics[height=0.12\textwidth, trim = 0 3mm 0 0, clip]{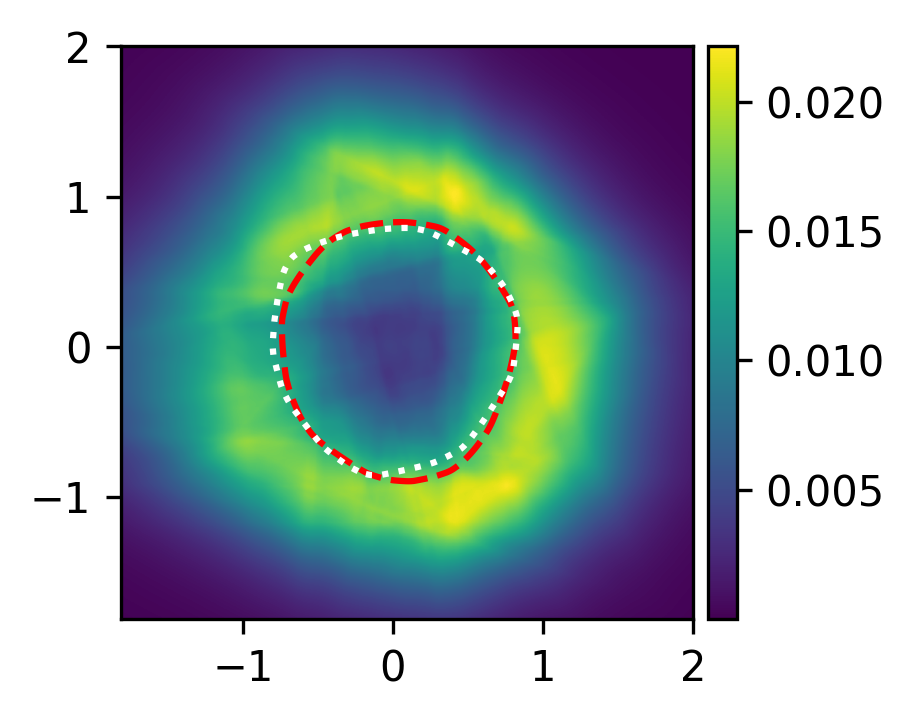} &
  \includegraphics[height=0.12\textwidth, trim = 0 3mm 0 0, clip]{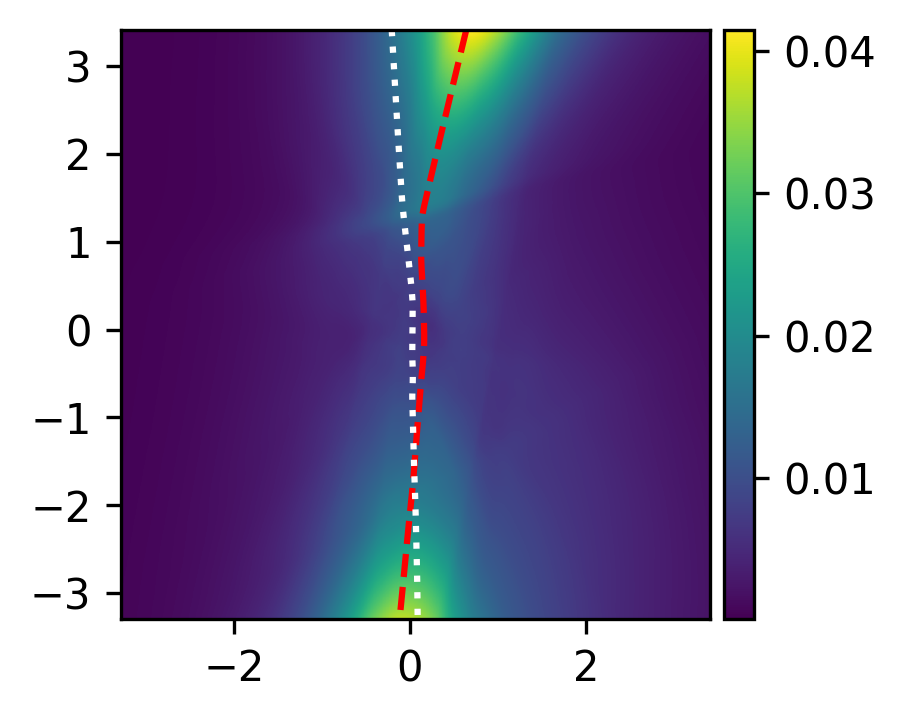} \\ \bottomrule
\rotatebox{90}{\quad \ DGE$_{20}$} &
  \includegraphics[height=0.12\textwidth, trim = 0 3mm 0 0, clip]{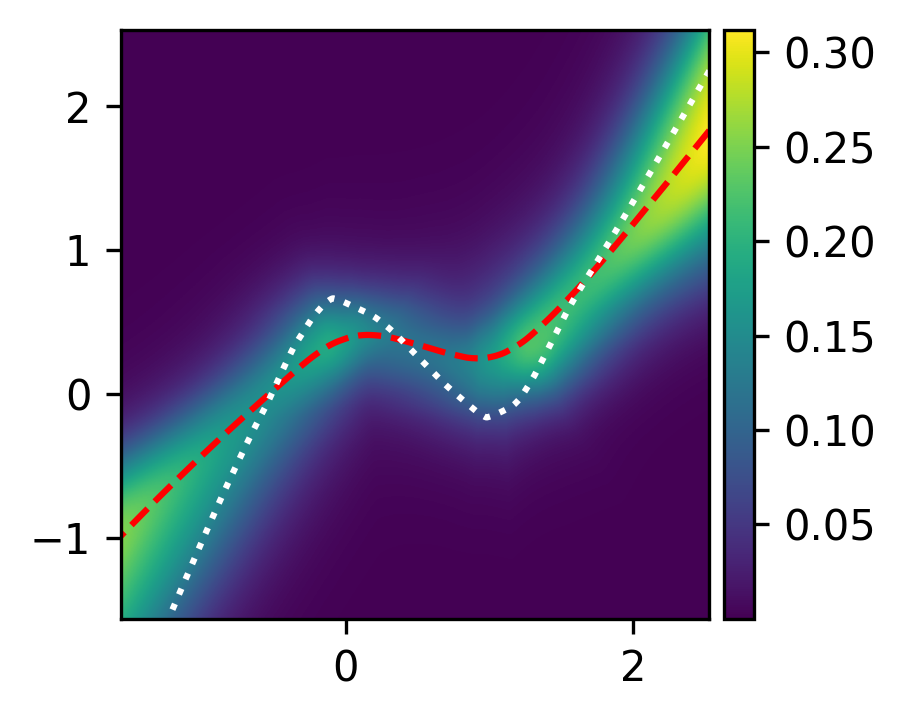} &
  \includegraphics[height=0.12\textwidth, trim = 0 3mm 0 0, clip]{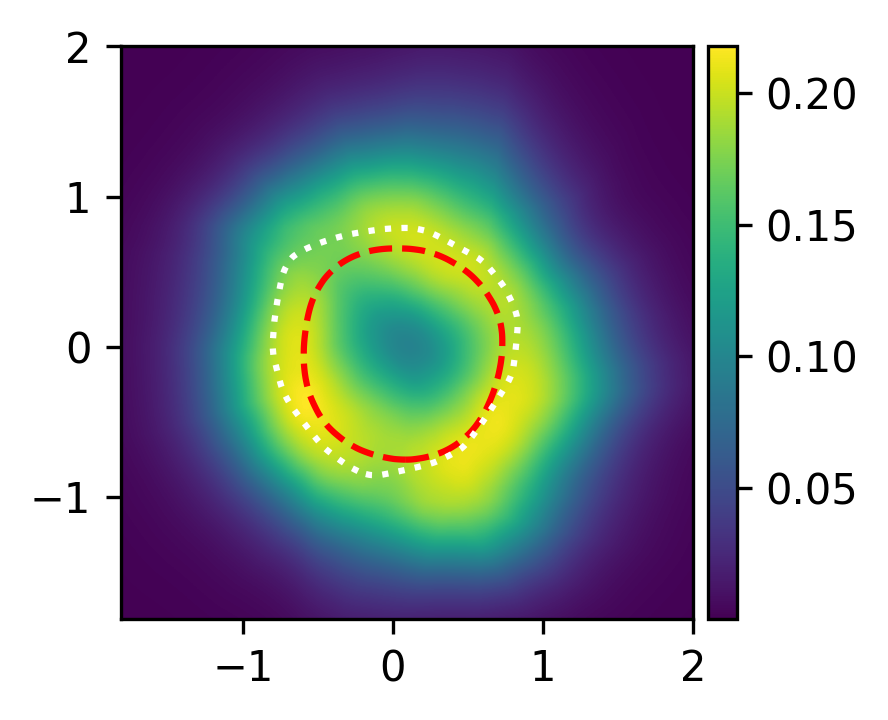} &
  \includegraphics[height=0.12\textwidth, trim = 0 3mm 0 0, clip]{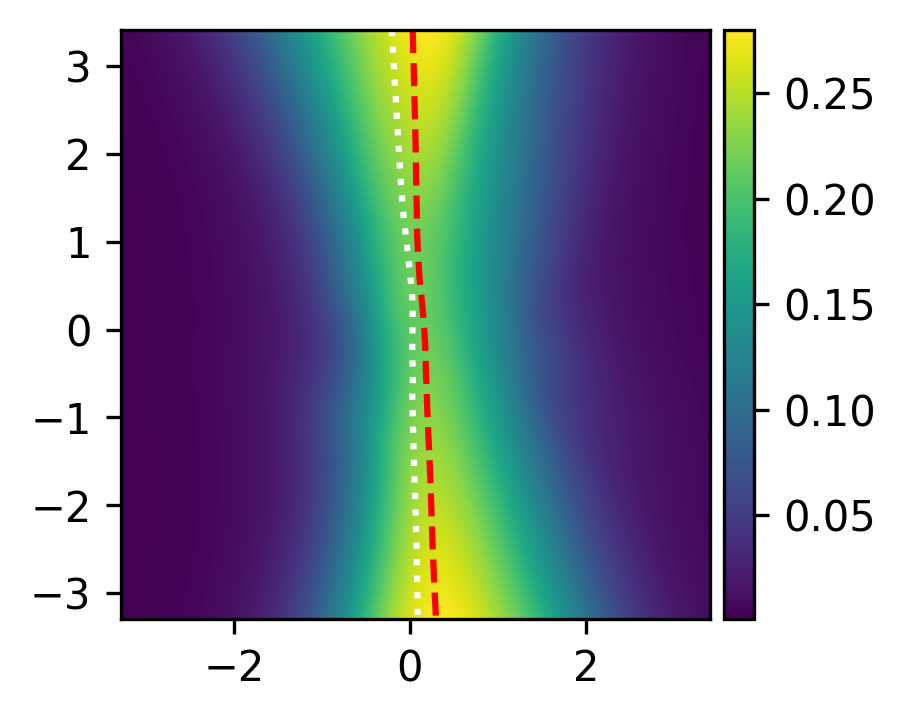} \\ \hline
\end{tabular}
    \caption{Comparison of predictive versus generative uncertainty. We plot the sample std of different ensembles, where columns denote datasets and rows approaches. The $\cD_r$ decision boundary ($\hat{P}(Y=1|x)=0.5$) is drawn in dotted white and other decision boundaries in dashed red. In almost all cases, these decision boundaries are significantly different. Meanwhile, the std of the naive approaches does not reflect this deviation, hence it underestimates the uncertainty---N.B. the different color scales. This is caused by these methods not considering the generative uncertainty. DGE$_{20}$ is preferred, as it does reflect generative uncertainty.}
    \label{fig:variance_plots}
\end{figure}


\textbf{Results.} First, let us draw the confidence-accuracy curves for different methods on the real-world datasets, see Figure \ref{fig:confidence_accuracy}. We see that DGE is able to make consistently more confident predictions more successfully than the naive approach. DGE performs en par with the $\cD_r$-model, and in some cases outperforms it; this is likely due to the generative models effectively augmenting the data, which has been shown to increase downstream robustness \citep{Antoniou2017DataNetworks, dina2022effect, das2022conditional, Bing2022ConditionalPopulations}.

Let us try to understand why uncertainty quantification on the single synthetic dataset does not suffice, by separating generative and predictive uncertainty. Specifically, let us plot the sample standard deviation of the different classifiers in the naive Deep Ensembles approach versus the DGE$_{20}$ approach, see Figure \ref{fig:variance_plots} and nota bene the different colorbar scales. We include the Naive (C) baseline, which is a naive approach that simply concatenates all synthetic datasets and runs a Deep Ensembles on this, but does not explicitly take into account generative uncertainty. We include this baseline to show that typical generative ensembles that result in a single dataset (see Table \ref{tab:related_work}), fail for UQ. We see that the naive approaches lead to poor diversity between different models within the ensemble. Since they cannot capture the generative uncertainty, these approaches provide little insight into how the model may do on real data. On the other hand, the diversity between DGE$_{20}$ classifiers is much higher and arguably provides more intuitive explanations. We also see that the naive approaches overestimate confidence in low-density regions---even if it is on a decision boundary---whereas DGE does not. This is unsurprising, since generative uncertainty is also highest in these regions.

\takeaway{DGE provides uncertainty on the generative level, which the naive approaches cannot. It is thus essential to ensure individual synthetic datasets in the DGE are published separately (cf. concatenated and shuffled)}

\subsection{Underrepresented Groups} \label{sec:underrepresented}
The generative process is expected to be most inaccurate for regions with few samples. Since low-density regions can correspond to minority groups in the population, this would be disconcerting for the reliability of our synthetic data. In this section we explore the quality of downstream models on underrepresented groups in the dataset.

\begin{figure}[hbt]
    \centering
    \includegraphics[width=\columnwidth]{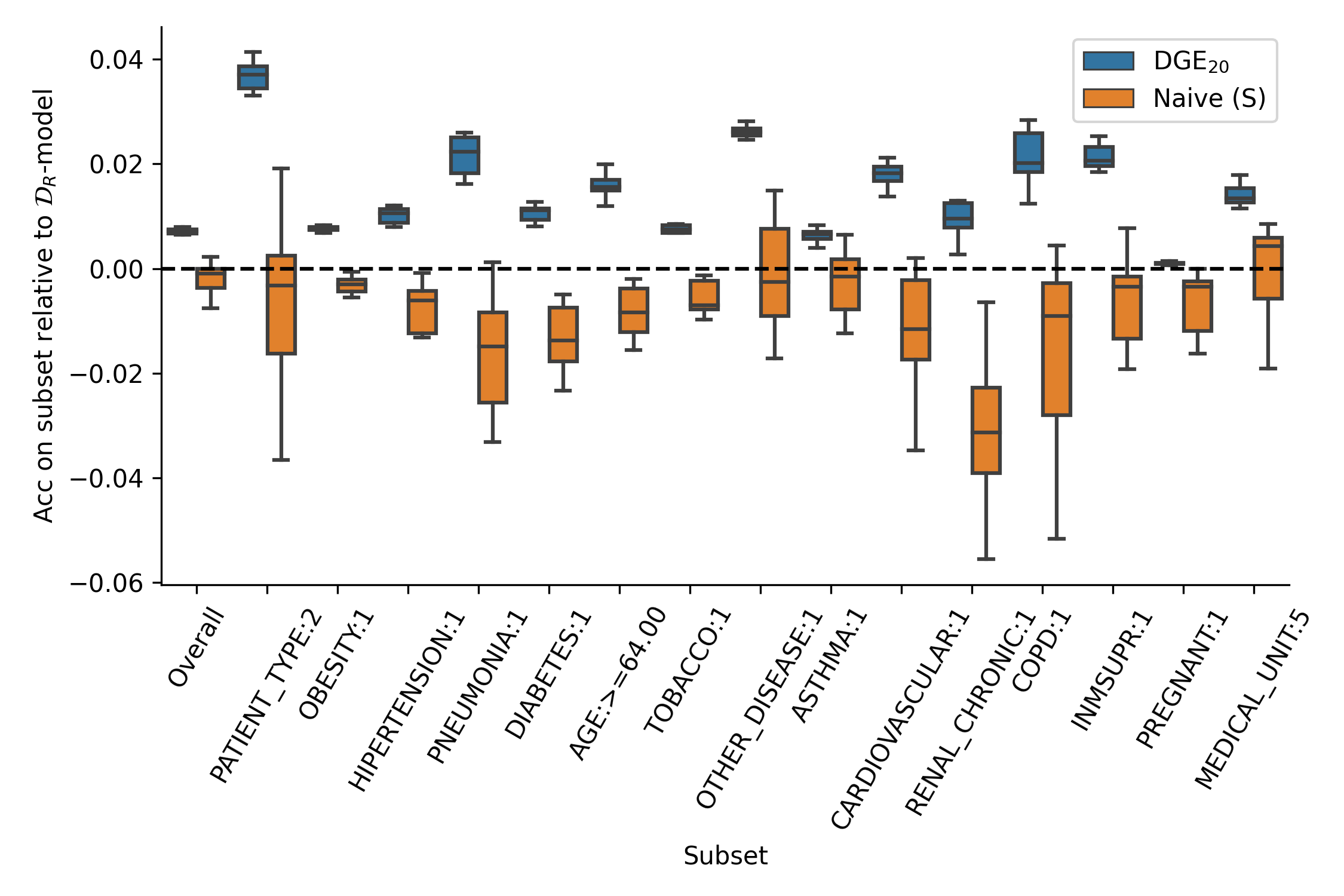}
    \caption{Accuracy of downstream model relative to $\cD_r$-model, evaluated on minority subsets. The naive approach tends to underperform the $\cD_r$-model on minority sets, whereas DGE outperforms the $\cD_r$-model.}
    \label{fig:underrepresented}
\end{figure}

\textbf{Set-up.} We investigate the Covid-19 dataset, because it consists of mostly categorical data with unbalanced features. Let us define ``underrepresented groups'' in terms of minority categories of individual features---see Appendix \ref{app:details}. We re-run the experiment from \ref{sec:training} and evaluate performance on the minority groups, see Figure \ref{fig:underrepresented}. We plot the performance relative to a $\cD_r$-model, which is trained on $\cD_r$ and also evaluated on underrepresented groups. 

\textbf{Results.} Note the distinctly different behaviour between the naive and DGE approach. The naive approach performs worse than the $\cD_r$-model for most underrepresented groups, even though it performs comparably overall. On the other hand, the DGE approach consistently outperforms the $\cD_r$-model. The latter is explained by interpreting DGE as a data augmentation method, in which the synthetic datasets (in this case $20$ times the size of the real data) replace the real data. Data augmentation can effectively regularize trained model \cite{Chawla2002SMOTE:Technique, Zhang2017Mixup:Minimization, Antoniou2017DataNetworks}, and lead to better performance on underrepresented groups \cite{Bing2022ConditionalPopulations}.

\takeaway{Closer inspection shows that naive downstream model training leads to particularly poor performance on small subgroups in the population. On the other hand, DGE has a regularization effect on the downstream predictor, consistently outperforming the $\cD_r$-model on minority groups.}

\section{Discussion} \label{sec:discussion}
\textbf{DGE.} We have aimed to highlight a gap in our understanding of synthetic data: how to do ML on synthetic data in practice, and the validity of downstream results. We have shown that the standard synthetic data methodology---treating the synthetic data like it is real---leads to poor downstream trained models, evaluation, and uncertainty, with a tendency to do worst for underrepresented groups. We have shown that DGE, which provides multiple synthetic datasets, is a simple and scalable way of avoiding these problems partly. We hope this work will result in a significant change in how synthetic datasets are published, create more interest in synthetic data's use and limitations, and contribute to the trustworthiness of synthetic data. Table \ref{tab:guidelines} highlights the takeaways for practitioners.

\begin{table}[bht]
    \centering 
    \caption{Guidelines for practitioners.}
    \label{tab:guidelines}
    {\color{white} \renewcommand{\arraystretch}{1.2}
    \begin{tabularx}{\columnwidth}{X}
    \cellcolor[HTML]{EA6B66} \makecell[c] {\textbf{Synthetic data publishers}} \\
          \cellcolor[HTML]{EA6B66} \textit{(1)} Generate multiple synthetic datasets with different random seeds. This allows better downstream ML, even if the number of generated datasets is small (e.g. $K=5$). \textit{(2)} These datasets need to be distinguishable---i.e. publishing each set in separate files or create a variable that denotes the set---since for some downstream tasks (e.g. UQ) it does not suffice to simply concatenate datasets without specifying the source of each sample. 
\textit{(3)} Include meta-data on the generative model, e.g. the amount of training data. Publishing the model class itself is also advisable, since the generative class can have implication on downstream model selection through congeniality. \\  \cellcolor[HTML]{7EA6E0}\makecell[c]{\textbf{Synthetic data users}} \\ \cellcolor[HTML]{7EA6E0}\textit{(1)} Do not treat synthetic data like it is real. \textit{(2)} Train models on an ensemble of synthetic datasets, to improve model performance on real data. \textit{(3)} For model evaluation and selection to reflect real-world performance, use train-test splits on the synthetic dataset level: train on some synthetic datasets, and evaluate on others. \textit{(4)} Be careful with UQ in the synthetic data regime, as naive UQ on a single dataset does not consider uncertainty in the generative process itself. By ensembling models trained on different synthetic datasets, one can quantify generative uncertainty. \textit{(5)} All of the above are extra relevant for ML on underrepresented groups, since the generative model is more likely to be inaccurate in these regions. 
\end{tabularx}}
\end{table}

\textbf{Practical Considerations.} Let us highlight practical considerations. First, DGE is not a perfect approximation for the true posterior of generative model parameters. Further research is required into systemic errors between synthetic datasets, e.g. if the true model is not well approximated by the generator class. Second, the use of ensembles requires extra compute at generation, downstream training, and downstream inference stage. In practice, we have seen that even for $K=5$ we can get significant gains compared to the naive approach. Additionally, cost may be further reduced by sharing parameters across models or using MC drop-out in the generative model. Third, each synthetic dataset is derived from the same real data, hence there is some data leakage between train and test sets. At last, if privacy is key, the ensemble approach of DGE requires scaling the privacy budget for each synthetic data generator to account for the multiple generators. Please see Appendix \ref{app:limitations} for a longer discussion.

\textbf{Extensions to Other Downstream Tasks.} We have explored using DGE for downstream prediction tasks, but future work could consider other domains (e.g. unsupervised learning, reinforcement learning, statistical analyses).


\section*{Acknowledgments}
We would like to thank the ICML reviewers and area chairs for their time and feedback, as well as Nabeel Seedat who reviewed an early draft of the paper. Additionally, we would like to acknowledge the Office of Navel Research UK, who funded this research.

\bibliography{mendeley, references_non_mendeley}
\bibliographystyle{icml2021}

\clearpage
\appendix
\section{Experimental Details}
\label{app:details}
\subsection{Data}
Throughout the paper we have used Scikit-learn's Two-moons and Circles toy datasets,  UCI's Breast Cancer and Adult Census Income \cite{Asuncion2007UCIRepository}, and SEER \cite{Duggan2016TheRelationship} and Mexican Covid-19 dataset \cite{MinistryofHealthofMexico2020Covid-19Dataset}, two large medical datasets with some highly unbalanced categorical features. For the uncertainty experiment, we have added a third toy dataset \textit{Gaussian}. The dataset was generated using:
\begin{align*}
    X_1, X_2 &\overset{iid}{\sim}\ N(0,1) \\
    Y &\sim Bern(t(X_1+1)) \\
    t(x) &= \begin{cases} 0 &, x<0\\
    1 &, x>2 \\
    \frac{1}{2}x &, otherwise
    \end{cases}
\end{align*}
This dataset was chosen because the optimal decision boundary is exact, $\hat{y}=1(x_1>0)$. 

\subsection{Experimental Settings}
Throughout, we have used CTGAN \cite{xu2019modeling} as the generative model, with hidden layer size 500 for both discriminator and generator. Unless otherwise stated, as downstream model we use Scikit-learn's MLP with 2 hidden layers of size 100, cross-entropy loss. We use learning rate $10^{-3}$, $L_2$ regularization $10^{-4}$, and optimization using Adam for all neural networks. 

\textbf{Sections \ref{sec:training}.} For CTGAN we use 2 hidden layers of sizes 500 for all datasets, and we use 2000 examples for training the generative model (except for breast cancer, for which we use 80\% of data) and generate 2000 samples.

The pipeline consists of the following.
\begin{enumerate}
    \item Real data is split into $\cD_{r,train}$ and $\cD_{r,test}$
    \item Generative model $G$ is trained on $\cD_{r,train}$ and used to generate. This is repeated $K=20$ times, for different random seeds.
    \item Each generative model is used to generate a respective dataset $\cD_{s}^k$.
    \item $\{\cD_{s}^k\}_{k=1}^K$ is used for training the different downstream approaches. \begin{enumerate}
        \item Naive single (S): the first synthetic dataset is used to train the downstream classifier.
        \item Naive ensemble (E): the first synthetic dataset is used to train $20$ classifiers with different seeds. Outputted probabilities are averaged.
        \item Naive concatenated (C): synthetic sets $\{\cD_{s}^k\}_{k=1}^20$ are concatenated and the full dataset is used for training downstream classifier.
        \item DGE$_{K}$: a classifier is trained on each of the synthetic datasets, and the resulting model is the ensemble that averages the outputted probabilities of the first $K$ classifiers.
        \item $\cD_r$-model: this model is trained on the real data, $\cD_{r,train}$, and \textbf{not} the synthetic data.
    \end{enumerate}
    \item Each approach's model is tested on $\cD_{r,test}$.
    \item Steps 2-5 are repeated for 10 runs with different seeds.
\end{enumerate}

\textbf{Section \ref{sec:evaluation}.} We increase the generative model and increase the number of training samples. Specifically, we use 5000 samples for training the generative model and generate 5000 samples. We use 3 hidden layers for CTGAN's discriminator and generator, and 2 hidden layers for the downstream model. In the Model Selection experiment, we use default settings for Scikit-learn classifiers---random forest has 100 estimators, kNN uses 5 neighbours, SVM uses an RBF kernel---and use 3 hidden layers for deep MLP and just 1 hiden layer for MLP.

The pipeline is similar as before, with the same steps 1-3.
\begin{enumerate}
    \item Real data is split into $\cD_{r,train}$ and $\cD_{r,test}$
    \item Generative model $G$ is trained on $\cD_{r,train}$ and used to generate. This is repeated $K=20$ times, for different random seeds.
    \item Each generative model is used to generate a respective dataset $\cD_{s}^k$.
    \item Each synthetic dataset is split into a train and test split $p_{train}=0.8$, $\cD_{s,train}^k$ and $\cD_{s,test}^k$.
    \item Each set $\cD_{s}^{train}$ is used for training a downstream model, $g_k$.
    \item $\{\cD_{s,test}^k\}_{k=1}^K$ are used for evaluating the models $g_k$, using different approaches. \begin{enumerate}
        \item Naive: $g_k$ is evaluated on $\cD^k_{S,test}$.
        \item DGE$_{K}$: $g_k$ is evaluated on $\cup_{i\neq k} \cD_{s,test}^{i}$ (limited to $K=5, 10$ or $20$ other datasets)
        \item Oracle: this is a pseudo-oracle that evaluates $g_k$ on real test set $\cD_r^{test}$.
    \end{enumerate}
    \item The previous step is repeated for $k=1, ..., 20$ and results are averaged per approach.
    \item (Model selection) For the model selection experiment, steps 4-7 are repeated for the aforementioned model classes (XGBoost, SVM, etc). The target ranking is given by the Oracle model.
\end{enumerate}

\textbf{Section \ref{sec:uncertainty}.}
We use a similar settings as in Section \ref{sec:training}---2000 generative training samples, two CTGAN hidden layers and one MLP hidden layer. The MLP uses cross-entropy, which is a proper scoring rule, hence this warrants its use for uncertainty quantification \cite{Lakshminarayanan2016SimpleEnsembles}. The pipeline is as follows.

\begin{enumerate}
    \item Real data is split into $\cD_{r,train}$ and $\cD_{r,test}$
    \item Generative model $G$ is trained on $\cD_{r,train}$ and used to generate. This is repeated $K=20$ times, for different random seeds.
    \item Each generative model is used to generate a respective dataset $\cD_{s}^k$.
    \item $\{\cD_{s}^k\}_{k=1}^K$ is used for training the different downstream Deep Ensembles approaches. \begin{enumerate}
        \item Naive ensemble (E): the first synthetic dataset is used to train $20$ classifiers with different seeds.
        \item Naive concatenated (C): synthetic sets $\{\cD_{s}^k\}_{k=1}^20$ are concatenated and the full dataset is used for training 20 downstream classifiers with different seeds.
        \item DGE$_{20}$: a classifier is trained on each of the synthetic datasets.
        \item $\cD_r$-model: 29 downstream models are trained on the real data, $\cD_{r,train}$.
    \end{enumerate}
    \item We take a grid of test data and create predictions for each approach.
    \item For each approach, we take the mean and standard deviation between members of the ensemble.
    \item We plot the contour where the mean equals 0.5 (i.e. the decision boundary), and plot the standard deviations over the full grid.
\end{enumerate}

\textbf{Section \ref{sec:underrepresented}.} We use the same set-up as Section \ref{sec:training}---2000 generative training samples with two hidden layers for both the CTGAN and the downstream MLP---however we use a different evaluation set (Step 5 in the pipeline). We take an agnostic approach in defining minorities: we loop through all features and for each feature we choose a minority subset based on this feature alone. For categorical features, we choose the smallest category with less than 20\% and more than 0.5\% of the population (and skip the feature if no such category exists). For continuous variable age, we choose the 10\% oldest patients. Categories are given in the figure behind the feature name.

\section{Type of Generative Model}
\label{app:gen_type}
In the main paper we have used CTGAN for all experiments, but in this section we show main results generalise to other deep generative models. We use the Synthcity library \cite{Qian2023Synthcity:Modalities} and generate data use different architectures, TVAE, a normalizing flow and ADS-GAN \cite{Yoon2020AnonymizationADS-GAN}---a privacy GAN. We use default settings for each. We repeat experiment \ref{sec:training}---see Table \ref{tab:other_generative_models}. We see the DGE$_K$ approaches perform significantly better on average, even for $K=5$.

\begin{table*}[hbtp]
\caption{Repeating experiment \ref{sec:training} for different generative model classes; AUC of downstream model trained on synthetic data, evaluated on real data. DGE$_K$ consistently outperforms the naive approaches on all datasets and generative model classes (except Moons for which it performs similarly).}
\label{tab:other_generative_models}
\begin{subtable}[b]{\textwidth}
\caption{TVAE}
\begin{tabular}{llllllll}
\toprule
{} &          Moons &        Circles &   Adult Income &  Breast Cancer &           SEER &       COVID-19 &   Mean \\
\midrule
Oracle              &    0.996 ± 0.0 &    0.868 ± 0.0 &    0.871 ± 0.0 &    0.993 ± 0.0 &    0.907 ± 0.0 &  0.928 ± 0.001 &  0.927 \\
Naive (S)           &  0.989 ± 0.003 &  0.856 ± 0.006 &  0.826 ± 0.012 &   0.97 ± 0.021 &  0.891 ± 0.005 &  0.867 ± 0.019 &    0.9 \\
Naive (E)           &  0.989 ± 0.003 &  0.856 ± 0.006 &   0.84 ± 0.009 &  0.979 ± 0.016 &  0.893 ± 0.005 &  0.892 ± 0.013 &  0.908 \\
Naive (C) &  0.991 ± 0.001 &  0.867 ± 0.002 &  0.854 ± 0.004 &  0.975 ± 0.011 &  0.907 ± 0.001 &  0.886 ± 0.005 &  0.913 \\
DGE$_{5}$           &  0.991 ± 0.001 &  0.866 ± 0.002 &  0.873 ± 0.003 &  0.984 ± 0.006 &  0.906 ± 0.001 &  0.917 ± 0.004 &  0.923 \\
DGE$_{10}$          &  0.991 ± 0.001 &  0.867 ± 0.001 &  0.883 ± 0.002 &  0.986 ± 0.002 &  0.908 ± 0.001 &  0.927 ± 0.002 &  0.927 \\
DGE$_{20}$          &    0.991 ± 0.0 &  0.868 ± 0.001 &   0.89 ± 0.001 &  0.987 ± 0.003 &    0.909 ± 0.0 &  0.934 ± 0.002 &   0.93 \\
\bottomrule
\end{tabular}
\end{subtable}
\begin{subtable}[b]{\textwidth}
\caption{ADS-GAN}
\begin{tabular}{llllllll}
\toprule
{} &          Moons &        Circles &   Adult Income &  Breast Cancer &           SEER &       COVID-19 &   Mean \\
\midrule
Oracle              &    0.996 ± 0.0 &    0.868 ± 0.0 &  0.871 ± 0.001 &    0.993 ± 0.0 &    0.907 ± 0.0 &  0.928 ± 0.001 &  0.927 \\
Naive (S)           &  0.981 ± 0.011 &  0.811 ± 0.042 &  0.811 ± 0.012 &   0.961 ± 0.02 &  0.886 ± 0.007 &  0.881 ± 0.021 &  0.888 \\
Naive (E)           &  0.981 ± 0.012 &   0.81 ± 0.043 &  0.829 ± 0.009 &  0.972 ± 0.012 &  0.888 ± 0.006 &  0.904 ± 0.012 &  0.897 \\
Naive (C) &  0.989 ± 0.001 &  0.861 ± 0.006 &  0.846 ± 0.003 &  0.965 ± 0.013 &  0.904 ± 0.001 &  0.897 ± 0.007 &   0.91 \\
DGE$_{5}$           &  0.988 ± 0.003 &  0.852 ± 0.011 &  0.872 ± 0.003 &  0.985 ± 0.006 &  0.902 ± 0.003 &  0.928 ± 0.006 &  0.921 \\
DGE$_{10}$          &  0.988 ± 0.002 &  0.863 ± 0.004 &  0.883 ± 0.002 &  0.986 ± 0.005 &  0.905 ± 0.001 &  0.937 ± 0.003 &  0.927 \\
DGE$_{20}$          &  0.988 ± 0.001 &  0.865 ± 0.002 &  0.889 ± 0.001 &  0.986 ± 0.003 &  0.906 ± 0.001 &  0.943 ± 0.002 &   0.93 \\
\bottomrule
\end{tabular}
\end{subtable}
\begin{subtable}[b]{\textwidth}
\caption{Normalizing flow}
\begin{tabular}{llllllll}
\toprule
{} &          Moons &        Circles &   Adult Income &  Breast Cancer &           SEER &       COVID-19 &   Mean \\
\midrule
Oracle              &    0.996 ± 0.0 &    0.868 ± 0.0 &    0.871 ± 0.0 &    0.993 ± 0.0 &    0.907 ± 0.0 &  0.928 ± 0.001 &  0.927 \\
Naive (S)           &  0.984 ± 0.007 &  0.778 ± 0.077 &  0.719 ± 0.041 &  0.966 ± 0.013 &   0.871 ± 0.02 &  0.595 ± 0.116 &  0.819 \\
Naive (E)           &  0.984 ± 0.007 &  0.781 ± 0.078 &  0.743 ± 0.038 &   0.971 ± 0.01 &  0.875 ± 0.018 &  0.678 ± 0.123 &  0.839 \\
Naive (C) &  0.986 ± 0.002 &  0.863 ± 0.002 &   0.773 ± 0.02 &   0.934 ± 0.02 &  0.898 ± 0.002 &  0.786 ± 0.039 &  0.873 \\
DGE$_{5}$           &  0.984 ± 0.005 &  0.857 ± 0.005 &   0.81 ± 0.012 &  0.978 ± 0.009 &  0.899 ± 0.003 &  0.852 ± 0.035 &  0.897 \\
DGE$_{10}$          &  0.984 ± 0.004 &  0.863 ± 0.003 &  0.845 ± 0.008 &  0.983 ± 0.005 &  0.903 ± 0.002 &  0.896 ± 0.019 &  0.912 \\
DGE$_{20}$          &  0.985 ± 0.001 &  0.865 ± 0.001 &  0.865 ± 0.006 &  0.984 ± 0.003 &  0.905 ± 0.001 &  0.923 ± 0.007 &  0.921 \\
\bottomrule
\end{tabular}
\end{subtable}
\end{table*}

\section{Effect of Synthetic Dataset Size}
\label{app:synthetic_dataset_size}
In the model selection experiments, we have seen that the naive approach's failure lies not only in the generative process; it is also dependent on the downstream model having the capacity to pick up on generative errors. We have seen that this implies that larger models are more likely to be chosen, since these may learn and copy overfitted correlations in the generated distribution best. In this appendix we explore another actor in the naive approach's failure: the synthetic dataset size. 

In Figure \ref{app:synthetic_dataset_size} we repeat experiment \ref{sec:selection}, but vary the synthetic dataset size from 500 to 20000. The oracle evaluation shows that the downstream trained models become slightly more stable (e.g. deep MLP) when more data is used, but their ranking does not change. The naive approach starts to significantly overestimate the complex models (random forest and deep MLP), giving a very poor and highly dataset-size variable ranking. The DGE approach to model evaluation and selection is much more stable, and follows the oracle closely.

\begin{figure*}[hbt]
    \centering
\begin{subfigure}[b]{0.33\textwidth}
\includegraphics[width=\textwidth]{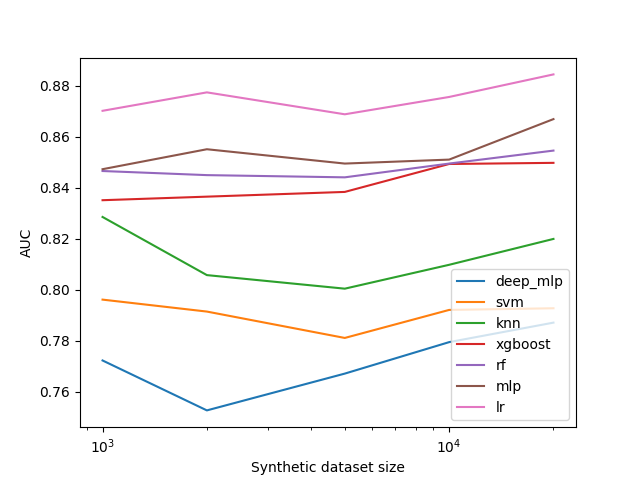}    
\caption{Oracle AUC}
\end{subfigure}
\begin{subfigure}[b]{0.33\textwidth}
\includegraphics[width=\textwidth]{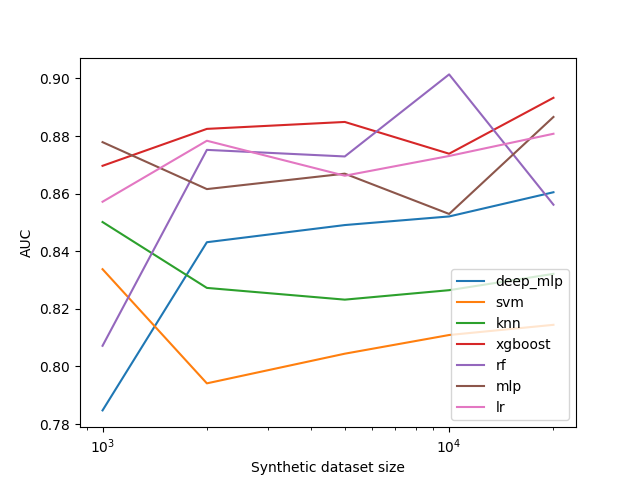}    
\caption{Naive AUC}
\end{subfigure}
\begin{subfigure}[b]{0.33\textwidth}
\includegraphics[width=\textwidth]{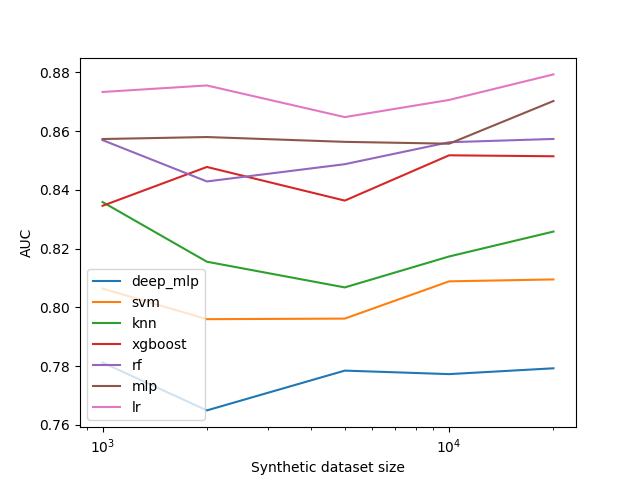}    
\caption{DGE$_{20}$ AUC}
\end{subfigure}
\begin{subfigure}[b]{0.33\textwidth}
\includegraphics[width=\textwidth]{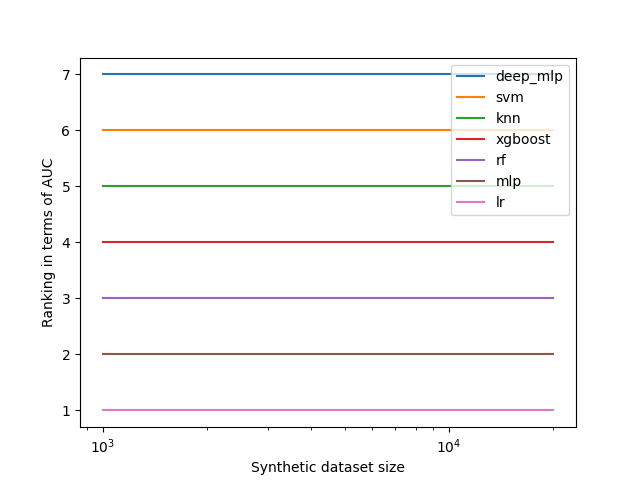}    
\caption{Oracle Ranking}
\end{subfigure}
\begin{subfigure}[b]{0.33\textwidth}
\includegraphics[width=\textwidth]{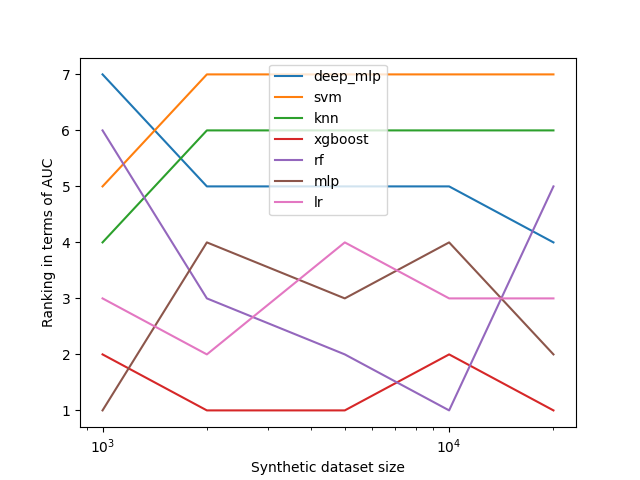}    
\caption{Naive Ranking}
\end{subfigure}
\begin{subfigure}[b]{0.33\textwidth}
\includegraphics[width=\textwidth]{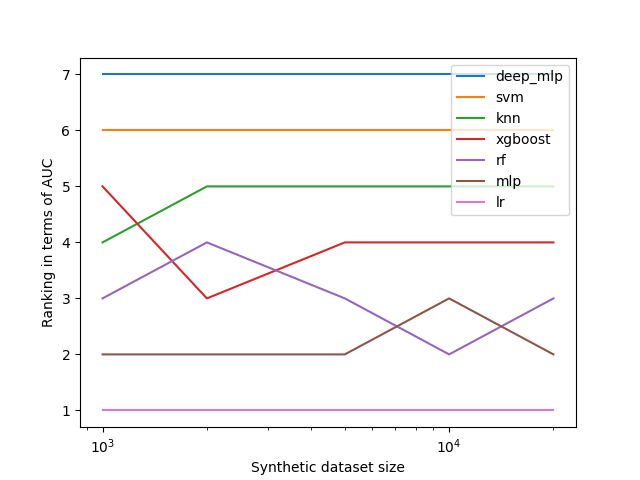}    
\caption{DGE$_{20}$ Ranking}
\end{subfigure}
\caption{\textbf{Varying synthetic dataset size has a large effect on the naive model evaluation and selection}. The naive approach starts overestimating complex models when the dataset size increases, while DGE$_{20}$ follows the oracle much more closely ($\pm 1$ difference). \textit{Ranking: lower is better}}    
    \label{fig:synthetic_dataset_size}
\end{figure*}


\section{Image Experiment}
\label{app:image}
In the main paper, we focused on tabular data--see Section \ref{sec:approximating posterior} for a motivation---and a GAN-based generator. Evidently, errors in the generative process are not constrained to tabular data. We include preliminary results for CIFAR-10 and CelebA (64$\times$ 64), generated using a small vanilla conditional Denoising Diffusion Probabilistic Model \cite{Ho2020DenoisingModels}. For both cases, we follow the same set-up as in Section \ref{sec:training}, though we use the full training split of the original data for training. For CelebA, we use annotation \textit{male} as label, and generate 10000 samples per class. For CIFAR-10, we generate 6000 samples per class. For the model trained on real data ($\mathcal{D}_r$), we use the full training split. We evaluate on the real test split. See results in Table \ref{tab:image}

\begin{table}[hbt]
    \caption{The advantage of DGE extends to the image domain. Even for small $K$, we can significantly outperform Naive (S) and (C).}
    \label{tab:image}
    \centering
\begin{tabular}{lll}
\toprule
Dataset &     CIFAR10 &CelebA \\
\midrule
$\mathcal{D}_r$ &  0.832$\pm$0.022 &  0.996$\pm$0.000 \\ \midrule
Naive (S)       &           0.662$\pm$0.013 &    0.8414$\pm$0.116\\
Naive (C)       &           0.690$\pm$0.013 &      0.887$\pm$0.054 \\ 
Naive (E)       &            0.716$\pm$0.004 &    \textbf{0.927$\pm$0.044} \\
DGE$_4$         &  \textbf{0.725$\pm$0.009} &     \textbf{0.927$\pm$0.045} \\
\bottomrule
    \end{tabular}
\end{table}

\section{Limitations} \label{app:limitations}
\textbf{Ensembles.} A limitation of DGE is the cost of training and inference scaling linearly with the number of models. In practice, this need not always be a problem. First, we have shown that even for $K=5$, synthetic datasets in the ensemble vastly outperform the currently-standard naive baseline. Second, generative models can be trained in parallel and so can downstream models. Third, the most expensive part of our experimentation pipeline is the generative stage, which in practice would be performed by a data publisher who runs DGE only once. Fourth, many datasets are not as big as high-resolution image datasets, and are thus cheaper to generate and train models on; training a CTGAN on the Adult dataset and generating data takes less than four minutes on a consumer PC with an RTX 3080 GPU. Fifth, though we propose a deep generative ensemble with unshared weights across generative models, this can be extended to partly shared weights—--e.g. future work could consider a single network with MC-dropout as UQ method \cite{Gal2015DropoutLearning}. 

Besides cost, a deep ensembles are a crude approximation for the true posterior of generative model parameters. The quality of DGE relies partly on how well the base model can approximate the true distribution. In the supervised learning literature, \citet{Wenzel2020HyperparameterQuantification} propose an ensemble method based on this idea, that ensembles not just over model parameters, but hyperparameters too. Other methods can be used for approximating the posterior over $\theta$. One can draw inspiration from UQ literature, e.g. use MC dropout UQ \cite{Gal2015DropoutLearning}, which often performs poorer than DE \cite{Fort2019DeepPerspective}, but scales better. \citet{Wen2020BatchEnsemble:Learning} aim to achieve the best of both worlds by sharing parameters across ensemble members, but promoting diversity in model outcome. Future work can extend these methods to the synthetic data generation regime. At last, generating multiple synthetic datasets has privacy implications. If differential privacy guarantees are required, one needs to scale the privacy budget of each synthetic dataset \cite{Dwork2006OurGeneration}.

\textbf{Data Leakage.} In this work we follow the usual Deep Ensembles approach and train each generative model on all real data. Though the original paper \cite{Lakshminarayanan2016SimpleEnsembles} finds that this performs better for uncertainty estimation, one may wonder whether this does not lead to an underestimate of generalisation error, due to the synthetic test and training sets both being derived from the same real data. This bias can be reduce by training the generators on disjoint subsets of the real data, such that one can train downstream models on some synthetic dataset, and evaluate on another. The downside of this naive approach is that the amount of training data for each generator is just $1/K$ the total amount, and that that by itself could lead to weaker (or more overfitted) generators. In this work we have decided to use all data for training the generators, following \cite{Lakshminarayanan2016SimpleEnsembles}. As seen in Section \ref{sec:evaluation}, there are no signs for a positive bias in model evaluation when using the DGE method, despite all generative models using the same training data. 
This is possibly explained the hope that generators may overfit to some data, but that different generators overfit in different ways---hence the variance across downstream results from different synthetic datasets is a correct measure of uncertainty. This could motivate increasing independence of generators, e.g. through different model choices and hyperparameters, but we leave this for further work.

\end{document}